\documentclass[10pt,twocolumn,letterpaper]{article}

\usepackage{cvpr}              

\usepackage[accsupp]{axessibility}  

\usepackage{graphicx}
\usepackage{amsmath,amssymb,mathtools,amsthm}
\usepackage{booktabs}
\usepackage{lipsum} 
\usepackage{multirow}
\usepackage{float} 
\usepackage{mathtools} 
\usepackage{algorithm,algorithmicx,algpseudocode}
\usepackage{color}
\usepackage[table]{xcolor}

\usepackage{placeins}
\usepackage[export]{adjustbox}
\usepackage{bbding}

\usepackage[accsupp]{axessibility}

\usepackage{pythonhighlight}

\usepackage{hyperref}
\hypersetup{colorlinks,allcolors=black}

\usepackage[capitalize]{cleveref}
\crefname{section}{Sec.}{Secs.}
\Crefname{section}{Section}{Sections}
\Crefname{table}{Table}{Tables}
\crefname{table}{Tab.}{Tabs.}

\def\cvprPaperID{1352} 
\def\confName{CVPR}
\def\confYear{2023}

\begin{document}

\title{LayoutDiffusion: Controllable Diffusion Model for \\ Layout-to-image Generation}


\newcommand*\samethanks[1][\value{footnote}]{\footnotemark[#1]}

\author{Guangcong Zheng$^1$\thanks{The first two authors contributed equally to this paper.} , Xianpan Zhou$^2$\samethanks[1] , Xuewei Li$^1$\thanks{Corresponding author.} , Zhongang Qi$^3$, Ying Shan$^3$, Xi Li$^{1,4,5,6}$\samethanks[2]
\\
$^1$College of Computer Science \& Technology, Zhejiang University
\\
$^2$Polytechnic Institute, Zhejiang University \enspace $^3$ARC Lab, Tencent PCG
\\
$^4$Shanghai Institute for Advanced Study of Zhejiang University 
\\
$^5$Shanghai AI Lab \enspace $^6$Zhejiang – Singapore Innovation and AI Joint Research Lab
\\
{\tt \small \{guangcongzheng, zhouxianpan, xueweili, xilizju\}@zju.edu.cn}
\\
{\tt \small \{zhongangqi, yingsshan\}@tencent.com}
}

\twocolumn[{%
\renewcommand\twocolumn[1][]{#1}%

\maketitle

\vspace{-2.5em}

\begin{center}
    \includegraphics[width=0.95\linewidth]{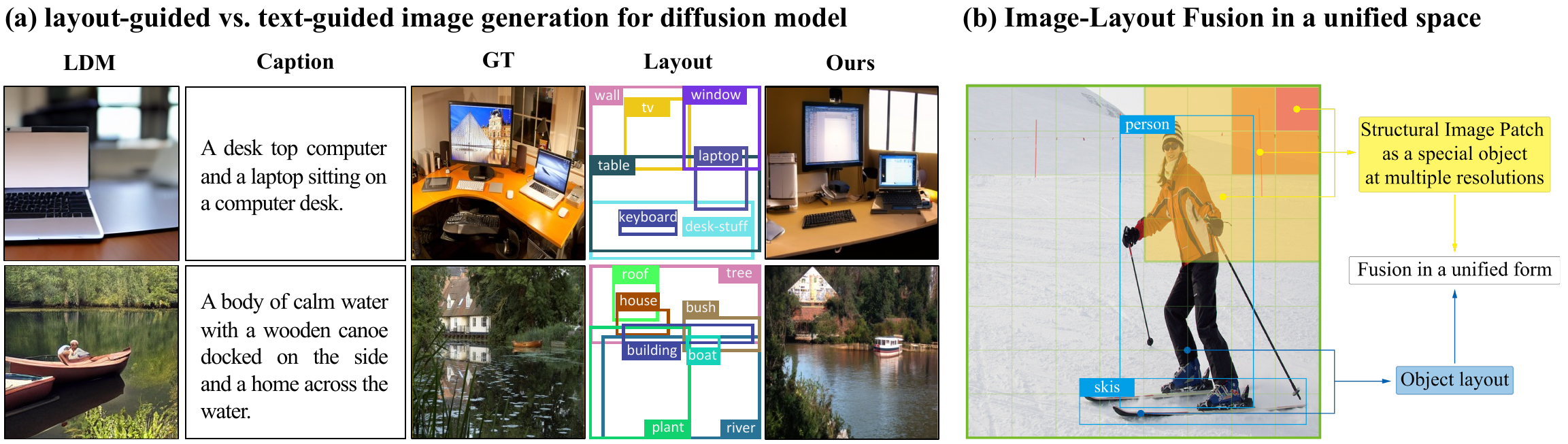}
    \captionof{figure}{
     Compared to text, the layout allows diffusion models to obtain more control over the objects while maintaining high quality. 
      Unlike the prevailing methods, we propose a diffusion model named LayoutDiffusion for layout-to-image generation. We transform the difficult multimodal fusion of the image and layout into a unified form by constructing a structural image patch with region information and regarding the patched image as a special layout.
    }
    \label{fig:first_img}
\end{center}
}]

\def\thefootnote{*}\footnotetext{Equal contribution.}
\def\thefootnote{$\dagger$}\footnotetext{Corresponding author.}

\hyphenation{LayoutDiffusion}

\begin{abstract}
Recently, diffusion models have achieved great success in image synthesis. However, when it comes to the layout-to-image generation where an image often has a complex scene of multiple objects, how to make strong control over both the global layout map and each detailed object remains a challenging task. In this paper, we propose a diffusion model named LayoutDiffusion that can obtain higher generation quality and greater controllability than the previous works. To overcome the difficult multimodal fusion of image and layout, we propose to construct a structural image patch with region information and transform the patched image into a special layout to fuse with the normal layout in a unified form. Moreover, Layout Fusion Module (LFM) and Object-aware Cross Attention (OaCA) are proposed to model the relationship among multiple objects and designed to be object-aware and position-sensitive, allowing for precisely controlling the spatial related information. Extensive experiments show that our LayoutDiffusion outperforms the previous SOTA methods on FID, CAS by relatively 46.35$\%$, 26.70$\%$ on COCO-stuff and 44.29$\%$, 41.82$\%$ on VG. Code is available at \url{https://github.com/ZGCTroy/LayoutDiffusion}.
\end{abstract}


\vspace{-1em}
\section{Introduction}
\label{sec:intro}

Recently, the diffusion model has achieved encouraging progress in conditional image generation, especially in text-to-image generation such as GLIDE~\cite{GLIDE}, Imagen~\cite{Imagen}, and Stable Diffusion~\cite{LDM}.
However, text-guided diffusion models may still fail in the following situations.
As shown in \cref{fig:first_img} (a), 
when aiming to generate a complex image with multiple objects, it is hard to design a prompt properly and comprehensively.
Even input with well-designed prompts, problems such as missing objects and incorrectly generating objects' positions, shapes, and categories still occur in the state-of-the-art text-guided diffusion model~\cite{GLIDE,LDM,Imagen}. 
This is mainly due to the ambiguity of the text and its weakness in precisely expressing the position of the image space~\cite{summaira2021recent, chen2014ranking, jiang2019learning, li2011graph, su2023language, su2023referring}. 
Fortunately, this is not a problem when using the coarse layout as guidance, which is a set of objects with the annotation of the bounding box (bbox) and object category. 
With both spatial and high-level semantic information, the diffusion model can obtain more powerful controllability while maintaining the high quality. 

However, early studies~\cite{sg2im, Grid2Im, LostGAN-v2, PLGAN} on layout-to-image generation are almost limited to generative adversarial networks (GANs) and often suffer from unstable convergence~\cite{arjovsky2017towards} and mode collapse~\cite{radford2015unsupervised}.
Despite the advantages of diffusion models in easy training~\cite{DDPM} and significant quality improvement~\cite{ADM-G},  few studies have considered applying diffusion in the layout-to-image generation task. 
To our knowledge, only LDM~\cite{LDM} supports the condition of layout and has shown encouraging progress in this field.

In this paper, different from LDM that applies the simple multimodal fusion method (e.g., the cross attention) or direct input concatenation for all conditional input, we aim to specifically design the fusion mechanism between layout and image. 
Moreover, instead of conditioning only in the second stage like LDM, we propose an end-to-end one-stage model that considers the condition for the whole process,  which may have the potential to help mitigate loss in the task that requires fine-grained accuracy in pixel space~\cite{LDM}.
The fusion between image and layout is a difficult multimodal fusion problem.
Compared to the fusion of text and image, the layout has more restrictions on the position, size, and category of objects.
This requires a higher controllability of the model and often leads to a decrease in the naturalness and diversity of the generated image. 
Furthermore, the layout is more sensitive to each token and the loss in token of layout will directly lead to the missing objects.

To address the problems mentioned above, we propose treating the patched image and the input layout in a unified form. Specifically, we construct a structural image patch at multi-resolution by adding the concept of region that contains information of position and size. 
As a result, each patch of the image is transformed into a special type of object, and the entire patched image will also be regarded as a layout. 
Finally, the difficult problem of multimodal fusion between image and layout will be transformed into a simple fusion with a unified form in the same spatial space of the image.  
We name our model LayoutDiffuison, a layout-conditional diffusion model with Layout Fusion Module (LFM), object-aware Cross Attention Mechanism (OaCA), and corresponding classifier-free training and sampling scheme.
In detail, LFM fuses the information of each object and models the relationship among multiple objects, providing a latent representation of the entire layout.
To make the model pay more attention to the information related to the object, we propose an object-aware fusion module named OaCA. Cross-attention is made between the image patch feature and layout in a unified coordinate space by representing the positions of both of them as bounding boxes. 
To further improve the user experience of LayoutDiffuison, we also make several optimizations on the speed of the classifier-free sampling process and could significantly outperform the SOTA models in 25 iterations.

Experiments are conducted on COCO-stuff~\cite{COCO-stuff} and Visual Genome (VG)~\cite{VG}. Various metrics ranging from quality, diversity, and controllability show that LayoutDiffusion significantly outperforms both state-of-the-art GAN-based and diffusion-based methods.

Our main contribution is listed below.

\begin{itemize}

\item Instead of using the dominated GAN-based methods, we propose a diffusion model named LayoutDiffusion for layout-to-image generations, which can generate images with both high-quality and diversity while maintaining precise control over the position and size of multiple objects.
 
\item We propose to treat each patch of the image as a special object and accomplish the difficult multimodal fusion of layout and image in a unified form. LFM and OaCA are then proposed to fuse the multi-resolution image patches with user's input layout.

\item LayoutDiffuison outperforms the SOTA layout-to-image generation method on FID, DS, CAS by relatively around 46.35$\%$, 9.61$\%$, 26.70$\%$ on COCO-stuff and 44.29$\%$, 11.30$\%$, 41.82$\%$ on VG.

\end{itemize}

\section{Related work}
\label{sec:related_work}
The related works are mainly from layout-to-image generation and diffusion models. 

\vspace{1em}
\noindent\textbf{Layout-to-Image Generation.}
Before the layout-to-image generation is formally proposed, the layout is usually used as as a complementary feature~\cite{karacan2016learning, reed2016learning, tan2018text2scene} or an intermediate representation in text-to-image~\cite{hong2018inferring}, scene-to-image generation~\cite{sg2im}.
The first image generation directly from the layout appears in Layout2Im~\cite{layout2im_conference} and is defined as a set of objects annotated with category and bbox.
Models that work well with fine-grained semantic maps at the pixel level can also be easily transformed to this setting~\cite{spade,pix2pix,hd-pix2pix}.
Inspired by StyleGAN~\cite{karras2019style}, LostGAN-v1~\cite{LostGAN-v1}, LostGAN-v2~\cite{LostGAN-v2} used a reconfigurable layout to obtain better control over individual objects. 
For interactive image synthesis, PLGAN~\cite{PLGAN} employed panoptic theory~\cite{kirillov2019panoptic} by constructing stuff and instance layouts into separate branches and proposed Instance- and Stuff-Aware Normalization to fuse into panoptic layouts. 
Despite encouraging progress in this field, almost all approaches are limited to the generative adversarial network (GAN) and may suffer from unstable convergence~\cite{arjovsky2017towards} and mode collapse~\cite{radford2015unsupervised}. 
As a multimodal diffusion model, LDM~\cite{LDM} supports the condition of coarse layout and has shown great potential in layout-guided image generation.
\paragraph{Diffusion Model.}
Diffusion models~\cite{dickstein, scorematching, DDPM, IDDPM, SDE, Analytic-DPM, LDM, yang2022diffusion_scene_graph} are being recognized as a promising family of generative models that have proven to be state-of-the-art sample quality for a variety of image generation benchmarks~\cite{croitoru2022diffusion,ulhaq2022efficient,yang2022diffusion}, including class-conditional image generation~\cite{ADM-G, ED-DPM}, text-to-image generation~\cite{GLIDE, LDM, Imagen}, and image-to-image translation~\cite{SR3, Repaint, DDRM}.
 Classifier guidance was introduced in ADM-G~\cite{ADM-G} to allow diffusion models to condition the class label. The gradient of the classifier trained on noised images could be added to the image during the sampling process. Then Ho et al.~\cite{classifier-free} proposed a classifier-free training and sampling strategy by interpolating between predictions of a diffusion model with and without condition input. For the acceleration of training and sampling speed, LDM proposed to first compress the image into smaller resolution and then apply denoising training in the latent space.

\begin{figure*}[ht]
  \centering
   \includegraphics[width=1.0\linewidth]{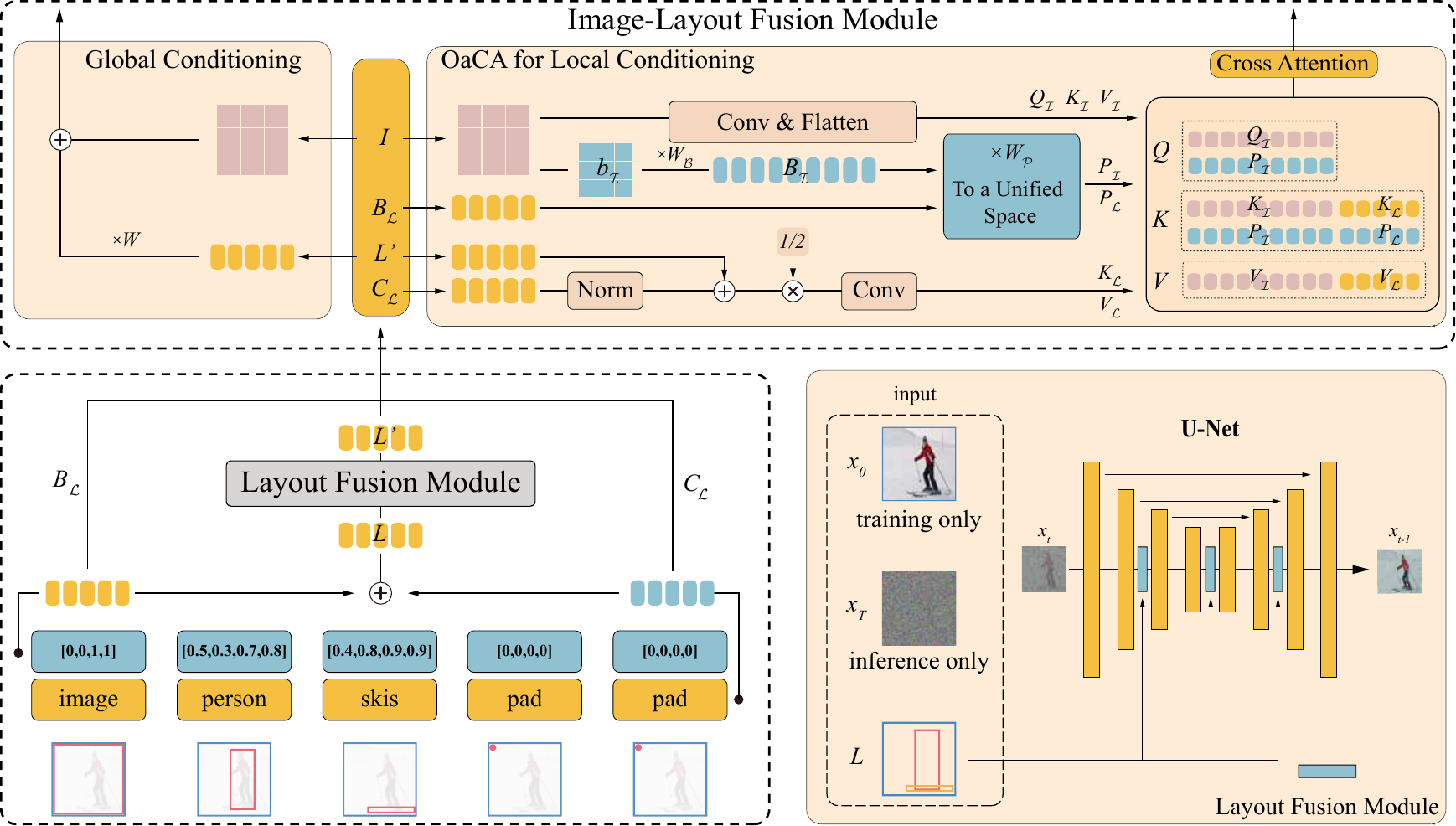}
   \caption{ The whole pipeline of LayoutDiffusion. The layout that consisted of bounding box $b$ and objects categories $c$ is transformed into embedding $B_{\mathcal{L}},C_{\mathcal{L}},L$. Then Layout Fusion Module fuses layout embedding $L$ to output the fused layout embedding $L'$. Finally, Image-Layout Fusion Module including direct addition used for global conditioning and Object-aware Cross Attention (OaCA) used for local conditioning, will fuse the layout related $B_{\mathcal{L}},C_{\mathcal{L}},L'$ and the image feature $I$ at multiple resolutions.}
   \label{fig:pipeline}
\end{figure*}


\section{Method}
\label{sec:method}


In this section, we propose our LayoutDiffusion, as shown in \cref{fig:pipeline}. The whole framework consists mainly of four parts: 
(a) layout embedding that preprocesses the layout input, 
(b) layout fusion module that encourages more interaction between objects of layout, 
(c) image-layout fusion module that constructs the structal image patch and object-aware cross attention developed with the specific design for layout and image fusion,
(d) the layout-conditional diffusion model with training and accelerated sampling methods.

\subsection{Layout Embedding}
\label{subsec:layout_input}
A layout $l \! = \! \{o_1, o_2, \cdots, o_n\}$ is a set of $n$ objects. Each object $o_i$ is represented as $o_i \! = \! \{b_i, c_i\}$, where $b_i \! = \! (x_0^i, y_0^i, x_1^i, y_1^i) \! \in \! [0,1]^4 $ denotes a bounding box (bbox) and $c_i \in [0, \mathcal{C}+1]$ is its category id.

To support the input of a variable length sequence, we need to pad $l$ to a fixed length $k$ by adding one 
$o_{l}$ in the front and some padding $o_{p}$ in the end, where $o_{l}$ represents the entire layout and $o_{p}$ represents no object. Specifically, $b_{l} \! = \! (0,0,1,1) $, $c_{l} \! =  \! 0$ denotes a object that covers the whole image and
$b_{p} \! = \! (0,0,0,0) $, $c_{p} \! =  \! \mathcal{C}+1$ denotes a empty object that has no shape or does not appear in the image. 

After the padding process, we can get a padded $l \! = \! \{o_1, o_2, \cdots, o_k\} \!$ consisting of $k$ objects, and each object has its specific position, size, and category. Then, the layout $l$ is transformed into a layout embedding $L \! = \! \{O_1, O_2, \cdots, O_k\} \! \in \! \mathbb{R}^{k \times d_{\mathcal{L}}}$ by the projection matrix $W_{\mathcal{B}} \in \mathbb{R}^{4 \times d_{\mathcal{L}}}$ and $W_\mathcal{C} \in \mathbb{R}^{1 \times d_{\mathcal{L}}}$ using the following equation:
\begin{align}
    L = & \, B_{\mathcal{L}} + C_{\mathcal{L}} \\
    B_{\mathcal{L}} &= b W_{\mathcal{B}} \label{eq:W_B}  \\
    C_{\mathcal{L}} &= c W_C
\end{align}
where $B_{\mathcal{L}}, C_{\mathcal{L}} \in \mathbb{R}^{k \times d_{\mathcal{L}}}$ are the bounding box embedding and the category embedding of a layout $l$, respectively. As a result, $L$ is defined as the sum of $B_{\mathcal{L}}$ and $C_{\mathcal{L}}$ to include both the content and positional information of a entire layout, and $d_{\mathcal{L}}$ is the dimension of the layout embedding.

\subsection{Layout Fusion Module}
\label{subsec:layout_fusion_module}

Currently, each object in layout has no relationship with other objects. This leads to a low understanding of the whole scene, especially when multiple objects overlap and block each other. Therefore, to encourage more interaction between multiple objects of the layout to better understand the entire layout before inputting the layout embedding, we propose Layout Fusion Module (LFM), a transformer encoder that uses multiple layers of self-attention to fuse the layout embedding and can be denoted as 
\begin{align}
    L' &= \operatorname{LFM}(L)
    \label{eq:layout_fusion_module}
\end{align}
, where the output is a fused layout embedding $ L' = \{O_{1}', O_{2}^{'}, \cdots, O_{k}^{'} \} \in \mathbb{R}^{k \times d_{\mathcal{L}}}$.

\subsection{Image-Layout Fusion Module}
\noindent\textbf{Structural Image Patch.}
The fusion of image and layout is a difficult multimodal fusion problem, and one of the most important parts lies in the fusion of position and size. However, the image patch is limited to the semantic information of the whole feature and lacks the spatial information. Therefore, we construct a structural image patch by adding the concept of region that contains the information of position and size.

Specifically, $I \in \mathbb{R}^{h \times w \times d_{\mathcal{I}}}$ denotes the feature map of a entire image with height $h$, width $w$, and channel $d_{\mathcal{I}}$. We define that $I_{u,v}$ is the $u^{\text{th}}$ row and $v^{\text{th}}$ column patch of $I$ and its bounding box, or the ablated region information, is defined as $b_{I_{u,v}}$ by the following equation:
\begin{align}
    b_{\mathcal{I}_{u,v}} & = (\frac{u}{h} \text{,} \frac{v}{w} \text{,} \frac{u\!+\!1}{h} \text{,} \frac{v\!+\!1}{w}) 
\end{align}
The bounding box sets of a patched image $I$ is defined as $b_{\mathcal{I}} \! = \! \{ b_{\mathcal{I}_{u,v}} | u \in [0,h), v \in [0, w)\}$. As a result, the positional information of image patch and layout object is contained in the unified bounding box defined in the same spatial space, leading to better fusion of image and layout.

\vspace{1em}
\noindent\textbf{Positional Embedding in Unified Space.}
We define the positional embedding of the image and layout as $P_{\mathcal{I}}$ and $P_{\mathcal{L}}$ as follows: 
\begin{align}
    B_{\mathcal{I}} & = b_{\mathcal{I}} W_{\mathcal{B}}  \\
    P_{\mathcal{I}} &= B_{\mathcal{I}} W_{\mathcal{P}} \\
    P_{\mathcal{L}} &= B_{\mathcal{L}} W_{\mathcal{P}}
\end{align}
, where $W_{\mathcal{B}} \! \in \! \mathbb{R}^{4 \times d_{\mathcal{L}}}$ is defined in Eq.~\ref{eq:W_B} and works as a shared projection matrix that transforms the coordinates of bounding box into embedding of $d_{\mathcal{L}}$ dimension. $W_{\mathcal{P}} \! \in \! \mathbb{R}^{d_{\mathcal{L}} \times d_{\mathcal{I}}}$ is the projection matrix that transforms the $B$ to the positional Embedding $P$.

\paragraph{Pointwise Addition for Global Conditioning.}
With the help of LFM in Eq.~\ref{eq:layout_fusion_module}, $O_{1}'$ can be considered as a global information of the entire layout, and $O_i' (i \in [2,k])$ 
is considered as the local information embedding of single object along with the other related objects. 
One of the easiest ways to condition the layout in the image is to directly add $O_{1}'$, the global information of the layout, to the multiple resolution of image features. Specifically, the condition process can be defined as
\begin{equation}
    I^{'} = I + O_{1}' W
\end{equation}
, where $W \in \mathbb{R}^{d_{\mathcal{L}} \times d_{\mathcal{I}}} $ is a projection matrix and $I'$ is the image feature conditioned with global embedding of layout.

\paragraph{Object-aware Cross Attention for Local Conditioning.}
Cross attention is successfully applied in ~\cite{GLIDE} to condition text into image feature, where the sequence of the image patch is used as the query and the concatenated sequence of the image patch and text is applied as key and value. The equation of cross-attention is defined as
\begin{equation}
\operatorname{Attention}(Q, K, V)=\operatorname{softmax}\left(\frac{Q {K}^T}{\sqrt{d_k}}\right) V
\label{eq:attention}
\end{equation}
, where $Q$, $K$, $V$ represent the embeddings of query, key, and value, respectively. In the following paper, we will use the subscript image and layout to represent the image patch feature and layout feature, respectively.

In text-to-image generation, each token in the text sequence is a word. The aggregation of these words constitutes the semantics of a sentence. After the transformer encoder, the first token in text sequence is well-semantic information that generalizes the whole text but may not reverse the semantic meaning of each word. However, the loss in information of one token is relatively serious in layout rather than in text. Each token in the layout sequence is a single object with a specific category, size, and position. The loss of information on a layout token will directly lead to a missing or wrong object in the generated image pixel space.

Therefore, we take into account the fusion of locations, size, and category of objects and define our object-aware cross-attention (OaCA) as
\begin{align}
     Q &= \operatorname{{\Psi}_{1}}(Q_{\mathcal{I}}, P_{\mathcal{L}}) \\
     K &= \operatorname{{\Psi}_{1}}(\operatorname{{\Psi}_{2}}(K_{\mathcal{I}}, K_{\mathcal{L}}), \operatorname{{\Psi}_{2}}(P_{\mathcal{I}}, P_{\mathcal{L}})) \\ 
     V &= \operatorname{{\Psi}_{2}}(V_{\mathcal{I}}, V_{\mathcal{L}})
    \label{eq:query_key_value}
\end{align}
, where the query $Q \in \mathbb{R}^{hw \times 2d_{\mathcal{I}}}$, $K \in \mathbb{R}^{(hw+k) \times 2d_{\mathcal{I}}}$, and $V \in \mathbb{R}^{(hw+k) \times d_{\mathcal{I}}} $. 
${\Psi}_{1}$ and $\Psi_{2}$ denote concatenation on the dimension of the channel and length of the sequence, respectively.

We first construct the key and value of the layout:
\begin{align}
    K_{\mathcal{L}}, V_{\mathcal{L}} &=  \operatorname{Conv}(\frac{1}{2}( \operatorname{Norm}(C_{\mathcal{L}}) + L')) 
    \label{eq:layout_key_value}
\end{align}
, where $K_{\mathcal{L}}, V_{\mathcal{L}} \in \mathbb{R}^{k \times d_{\mathcal{I}}}$ and $\operatorname{Conv}$ is the convolution operation. 
The embedding of key and value in the layout is related to the category embedding $C_{\mathcal{L}}$ and the fused layout embedding $L'$. 
$C_{\mathcal{L}}$ focuses on the category information of layout and  
$L'$ concentrates on the comprehensive information of both the object itself and other objects that may have a relationship with it. 
By averaging between $L'$ and $C_{\mathcal{L}}$, we can obtain both the general information of the object and also emphasize the category information of the object.

We construct the query, key, and value of the image feature as follows:
\begin{align}
    Q_{\mathcal{I}}, K_{\mathcal{I}}, V_{\mathcal{I}} =\operatorname{Conv}(\operatorname{Norm}(I)) 
    \label{eq:image_query_key_value}
\end{align}

\subsection{Layout-conditional Diffusion Model}
 Here, we follow the Gaussian diffusion models improved by~\cite{scorematching,DDPM}. 
 Given a data point sampled from a real data distribution $x_0 \sim q(x_0)$, a forward diffusion process is defined by adding small amount of Gaussian noise to the $x_0$ in $T$ steps:
\begin{align}
    q(x_t | x_{t-1}) \coloneqq \mathcal{N}(x_t; \sqrt{\alpha_t} x_{t-1}, (1-\alpha_t)\mathbf{I})
\end{align}
If the total noise added throughout the Markov chain is large enough, the $x_T$ will be well approximated by $\mathcal{N}(0, \mathbf{I})$. If we add noise at each step with a sufficiently small magnitude $1-\alpha_t$, the posterior $q(x_{t-1} | x_{t})$ will be well approximated by a diagonal Gaussian. This nice property ensures that we can reverse the above forward process and sample from $x_T \sim \mathcal{N}(0, \mathbf{I})$, which is a Gaussian noise. However, since the entire dataset is needed, we are unable to easily estimate the posterior. Instead, we have to learn a model $p_\theta(x_{t-1} | x_t)$ to approximate it:
\begin{align}
    p_{\theta}(x_{t-1}|x_t) \coloneqq \mathcal{N}(\mu_{\theta}(x_t), \Sigma_{\theta}(x_t))
\end{align}
Instead of using the tractable variational lower bound (VLB) in $\log p_\theta(x_0)$, Ho et al.\cite{DDPM} proposed to reweight the terms of the VLB to optimize a surrogate objective. Specifically, we first add $t$ steps of Gaussian noise to a clean sample $x_0$ to generate a noised sample $x_t \sim q(x_t | x_0)$. Then train a model $\epsilon_{\theta}$ to predict the added noise using the following loss:
\begin{equation}
    \mathcal{L} \! \coloneqq \! E_{t \sim [1,T],x_0 \sim q(x_0), \epsilon \sim \mathcal{N}(0, \mathbf{I})}[||\epsilon - \epsilon_{\theta}(x_t, t)||^2]
    \label{eq:lsimple}
\end{equation}
, which is a standard mean-squared error loss.

To support the layout condition, we apply classifier-free guidance, a technique proposed by Ho et al.~\cite{classifier-free} for conditional generation that requires no additional training of the classifier.
It is accomplished by interpolating between predictions of a diffusion model with and without condition input. For the condition of layout, we first construct a padding layout $l_{\phi} = \{o_{l}, o_{p}, \cdots, o_{p}\}$. 
During training, the condition of layout $l$ of diffusion model will be replaced with $l_{\phi}$ with a fixed probability. When sampling, the following equation is used to sample a layout-condional image:
\begin{equation}
    \hat{\epsilon}_{\theta}(x_t, t | l) = (1-s) \! \cdot \! \epsilon_{\theta}(x_t, t | l_{\phi}) + s \! \cdot \! \epsilon_{\theta}(x_t, t|l)
\end{equation}
, where the scale $s$ can be used to increase the gap between $\epsilon_{\theta}(x_t, t|l_{\phi})$ and $\epsilon_{\theta}(x_t, t|l)$ to enhance the strength of conditional guidance.

To further improve the user experience of LayoutDiffuison, we also make several optimizations on the speed of the classifier-free sampling process and could significantly outperform the SOTA models in 25 iterations. Specifically, we adapt DPM-solver~\cite{DPM-Solver} for the conditional classifier-free sampling, a fast dedicated high-order solver for diffusion ODEs~\cite{SDE} with the convergence order guarantee, to accelerate the conditional sampling speed.


\begin{figure*}[h!]
  \centering

   \includegraphics[width=1.0\linewidth]{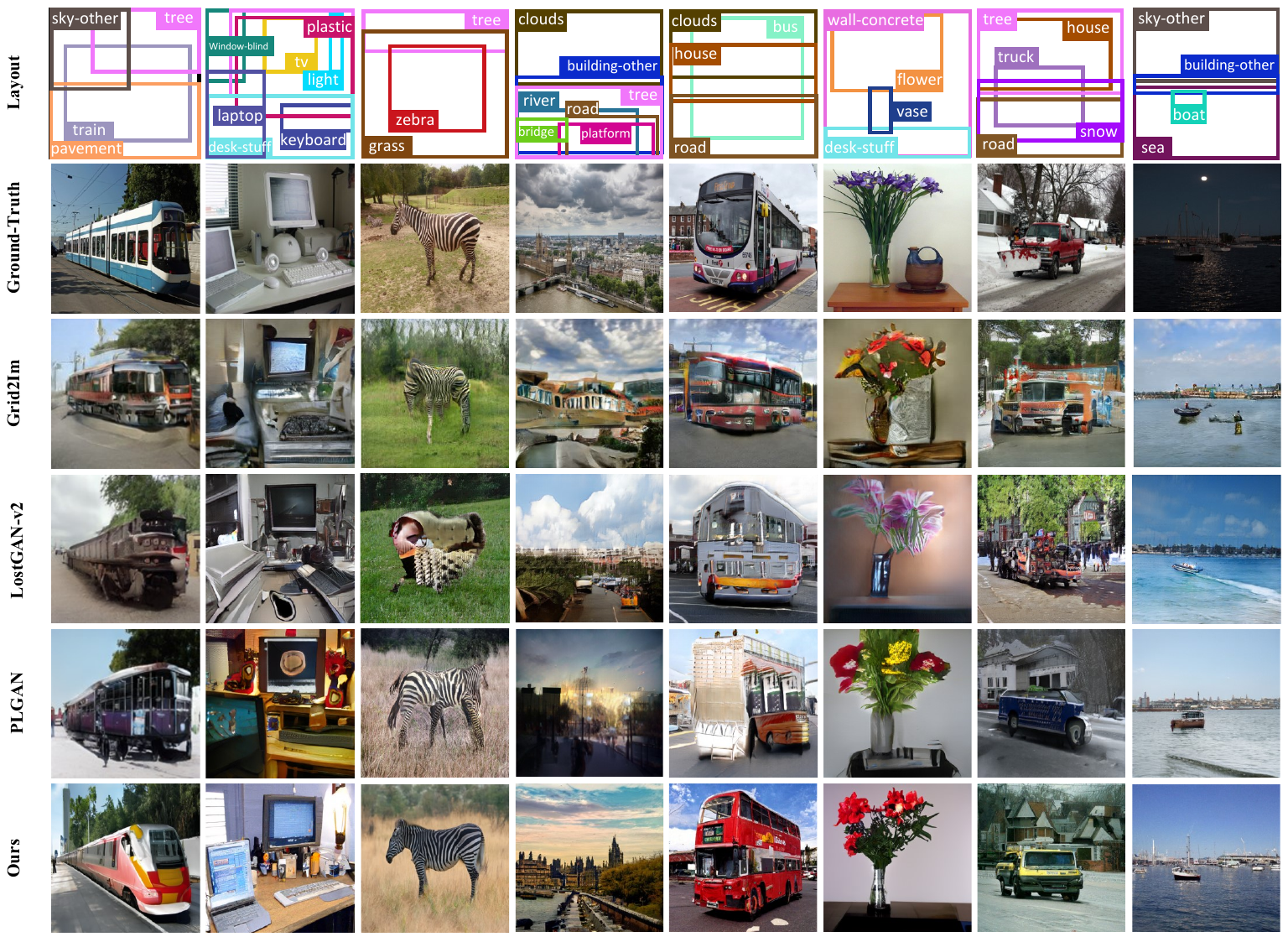}

   \caption{Visualization of comparision with SOTA methods on COCO-stuff 256$\times$256. 
   LayoutDiffusion has better generation quality and stronger controllability compared to the other methods.
   }
   \label{fig:visualization_of_comparision_with_SOTA}
\end{figure*}

\begin{table*}[h!]
  \centering
  \resizebox{1 \textwidth}{!}{
  \begin{tabular}{l|ccccc ccc}  
    \toprule
    \multirow{2}{4em}{\textbf{Methods}} & \multicolumn{5}{c}{\textbf{COCO-stuff}} & \multicolumn{3}{c}{\textbf{VG}} \\ 

    \cmidrule(lr){2-6} \cmidrule(lr){7-9}
    
     & \textbf{FID $\downarrow$} & \textbf{IS $\uparrow$}& \textbf{DS $\uparrow$} & \textbf{CAS $\uparrow$} & \textbf{YOLOScore $\uparrow$} & \textbf{FID $\downarrow$} & \textbf{DS $\uparrow$} & \textbf{CAS $\uparrow$} \\

    \midrule
    \rowcolor{gray!20}\textbf{128$\times$128}       &       &           &           &       &      &      &       &     \\  
    

    Grid2Im~\cite{Grid2Im}        & 59.50 & 12.50±0.30  & 0.28±0.11 & 4.05   & 6.80  & -     & -         & -     \\
    LostGAN-v2~\cite{LostGAN-v2}  & 24.76 & 14.21±0.40  & 0.45±0.09 & 39.91  & 13.60  & 29.00 & 0.42±0.09 & 29.74  \\
    PLGAN~\cite{PLGAN}            & 22.70 & 15.60±0.30  & 0.16±0.72 & 38.70  & 13.40  & 20.62 & -         & -     \\
    \textbf{LayoutDiffusion}      & \textbf{16.57} & \textbf{20.17±0.56}  & \textbf{0.47±0.09} & \textbf{43.60}  & \textbf{27.00} & \textbf{16.35} & \textbf{0.49±0.09} & \textbf{36.45}  \\

    \midrule
    \rowcolor{gray!20}\textbf{256$\times$256}       &       &      &           &       &      &      &       &   \\

    Grid2Im~\cite{Grid2Im}        & 65.20 & 16.40±0.70  & 0.34±0.13 & 4.81   & 9.70  & -     & -         & -     \\
    LostGAN-v2~\cite{LostGAN-v2}  & 31.18 & 18.01±0.50  & 0.56±0.10 & 40.00   & 17.50 & 32.08 & 0.53±0.10 & 34.48   \\
    PLGAN~\cite{PLGAN}            & 29.10 & 18.90±0.30  & 0.52±0.10 & 37.65  & 14.40  & 28.06 & -   & -     \\
    \textbf{LayoutDiffusion}      & \textbf{15.61} & \textbf{28.36±0.75}  & \textbf{0.57±0.10} & \textbf{47.74}   & \textbf{32.00}  & \textbf{15.63}  & \textbf{0.59±0.10} & \textbf{48.90} \\

    \bottomrule
    
  \end{tabular}
  }
  \caption{Quantitative results on COCO-stuff~\cite{COCO-stuff} and VG~\cite{VG}. 
  The proposed diffusion method has made great progress in all evaluation metrics, showing better quality, controllability, diversity, and accuracy than previous works. 
  For COCO-stuff, we evaluate on 3097 layout and sample 5 images for each layout. For VG, we evaluate on 5096 layout and sample 1 image for each layout.
  We also report reproduction scores of previous works in Appendix.
  } 
  \label{tab:quantitative_results1}
\end{table*}

\begin{table*}[htb]
  \centering
  \vspace{-0.5em}
  \resizebox{0.9 \textwidth}{!}{
  \begin{tabular}{@{}ccc|cccccc@{}}
    \toprule
    \textbf{LFM} & \textbf{OaCA} & \textbf{CA} &  FID $\downarrow$ & IS $\uparrow$  & DS $\uparrow$  & CAS $\uparrow$ & YOLOScore $\uparrow$\\ 
    \midrule
               &             &             & 29.94 \textcolor[RGB]{91, 179, 24}{(+13.37)}  & 13.59±0.29 \textcolor[RGB]{91, 179, 24}{(-6.58)}   & \textbf{0.70±0.08} \textcolor{red}{(+0.23)}   & 3.83 \textcolor[RGB]{91, 179, 24}{(-39.77)} & 0.00 \textcolor[RGB]{91, 179, 24}{(-27.00)}\\

    \checkmark &             &             & 17.06 \textcolor[RGB]{91, 179, 24}{(+0.49)}  & 19.21±0.53 \textcolor[RGB]{91, 179, 24}{(-0.96)}& 0.52±0.09 \textcolor{red}{(+0.05)} & 30.86 \textcolor[RGB]{91, 179, 24}{(-12.74)} & 6.90 \textcolor[RGB]{91, 179, 24}{(-20.10)} \\
               
               &  \checkmark &             & 16.76 \textcolor[RGB]{91, 179, 24}{(+0.19)}  & 19.57±0.40 \textcolor[RGB]{91, 179, 24}{(-0.60)} & 0.48±0.09 \textcolor{red}{(+0.01)} & 40.67 \textcolor[RGB]{91, 179, 24}{(-2.93)} & 18.80 \textcolor[RGB]{91, 179, 24}{(-8.20)}  \\
    
    \checkmark &             & \checkmark  & \textbf{16.46} \textcolor{red}{(-0.11)} & 19.79±0.40 \textcolor[RGB]{91, 179, 24}{(-0.38)} & 0.48±0.10 \textcolor{red}{(+0.01)}  & 42.47 \textcolor[RGB]{91, 179, 24}{(-1.13)} & 23.60 \textcolor[RGB]{91, 179, 24}{(-3.40)}  \\
    
    \checkmark &  \checkmark &             & 16.57  & \textbf{20.17±0.56} & 0.47±0.09 & \textbf{43.60} & \textbf{27.00}  \\

    \bottomrule
  \end{tabular}
  }
  \vspace{-0.4em}
  \caption{Ablation study of \textbf{Layout Fusion Module (LFM)}, \textbf{Object-aware Cross Attention (OaCA)}, \textbf{Cross Attention (CA)}. We use the model trained for 300,000 iterations on COCO-stuff 128$\times$128. The value in brackets denotes the discrepancy to our proposed method(+LFM+OaCA), where red denotes better and green denotes worse.
  }
  \label{tab:ablation_study}
  \vspace{-1.5em}
\end{table*}


\section{Experiments}
\label{sec:experiments}

In this section, we evaluate our LayoutDiffusion on different benchmarks in terms of various metrics. 
First, we introduce the datasets and evaluation metrics.
Second, we show the qualitative and quantitative results compared with other strategies. 
Finally, some ablation studies and analysis are also mentioned. 
More details can be found in Appendix, including model architecture, training hyperparameters, reproduction results, more experimental results and visualizations.

\subsection{Datasets}
\label{subsec:datasets}
We conduct our experiments on two popular datasets, COCO-Stuff~\cite{COCO-stuff}  and Visual Genome~\cite{VG}.

\vspace{0.5em}
\noindent\textbf{COCO-Stuff}
has 164K images from COCO 2017, of which the images contain bounding boxes and pixel-level segmentation masks for 80 categories of thing and 91 categories of stuff, respectively.
Following the settings of LostGAN-v2~\cite{LostGAN-v2}, we use the COCO 2017 Stuff Segmentation Challenge subset that contains 40K / 5k / 5k images for train / val / test-dev set, respectively. 
We use images in the train and val set with 3 to 8 objects that cover more than $2\%$ of the image and not belong to \textit{crowd}.
Finally, there are 25,210 train and 3,097 val images.

\vspace{0.5em}
\noindent\textbf{Visual Genome}
collects 108,077 images with dense annotations of objects, attributes, and relationships. 
Following the setting of SG2Im~\cite{sg2im}, we divide the data into 80$\%$, 10$\%$, 10$\%$ for the train, val, test set, respectively. 
We select the object and relationship categories occurring at least 2000 and 500 times in the train set, respectively, and select the images with 3 to 30 bounding boxes and ignoring all small objects. 
Finally, the training / validation / test set will have 62565 / 5062 / 5096 images, respectively.

\subsection{Evaluation Metrics \& Protocols}
\label{subsec:evaluation_metrics}

We use five metrics to evaluate the quality, diversity, and controllability of generation.

\noindent\textbf{Fr`echet Inception Distance (FID)}~\cite{FID}
shows the overall visual quality of the generated image by measuring the difference in the distribution of features between the real images and the generated images on an ImageNet-pretrained Inception-V3\cite{Inceptionv3} network.

\noindent\textbf{Inception Score (IS)}~\cite{IS} 
uses an Inception-V3\cite{Inceptionv3} pretrained on ImageNet network to compute the statistical score of the output of the generated images.

\noindent\textbf{Diversity Score (DS)}
calculates the diversity between two generated images of the same layout by comparing the LPIPS~\cite{LPIPS} metric in a DNN feature space between them.

\noindent\textbf{Classification Score (CAS)}~\cite{CAS} first crops the ground truth box area of images and resizing them at a resolution of 32$\times$32 with their class. 
A ResNet-101~\cite{ResNet} classifier is trained with generated images and tested on real images.

\noindent\textbf{YOLOScore}~\cite{YOLOScore}
evaluates 80 thing categories bbox mAP on generated images using a pretrained YOLOv4~\cite{YOLOv4} model, and shows the precision of control in one generated model. 

In summary, FID and IS show the generation quality, DS shows the diversity, CAS and YOLOScore represent the controllability. We follow the architecture of ADM~\cite{ADM-G}, which is mainly a UNet. All experiments are conducted on 32 NVIDIA 3090s with mixed precision training~\cite{micikevicius2017mixed}. We set batch size 24, learning rate 1e-5. We adopt the fixed linear variance schedule. More details can be found in the Appendix.


\subsection{Qualitative results}
\label{subsec:qualitative_results}
Comparison of generated 256 $\times$ 256 images on the COCO-Stuff~\cite{COCO-stuff} with our method and previous works~\cite{Grid2Im, LostGAN-v2, PLGAN} is shown in~\cref{fig:visualization_of_comparision_with_SOTA}.
\begin{table}[h!]
  \centering
  \vspace{-0.8em}
    \resizebox{0.48 \textwidth}{!}{
  \begin{tabular}{l|cccccc}
    \toprule
    
     \multirow{2}{*}{method} & FID $\downarrow$ & $N_{\text{parms}}$  & Throughout    & First stage   & Cond stage    \\
                             &                  &                    &  images / s   & V100 days & V100 
days \\ 
    \midrule
     LDM-8  (100 steps) &   42.06 &  345M  &   0.457   & 66 &   3.69     \\
     LDM-4  (200 steps) & 40.91  &  306M   &   0.267   & 29 &  95.49   \\
     ours-small (25 steps)              & 36.16    &  142M   &   0.608   & - & 75.83 \\ 
     ours  (25 steps)                 &    31.68   &   569M    &   0.308   & - & 216.55  \\
    \bottomrule
  \end{tabular}
  }
  \vspace{-0.5em}
  \caption{Comparison  with SOTA diffusion-based methods LDM on COCO-stuff 256$\times$256. We generate the same 2048 images of LDM for a fair comparision. }
  \vspace{-1.2em}
  \label{tab:ablation_study2}
\end{table}

LayoutDiffusion generates more accurate high quality images, which has more recognizable and accurate objects corresponding to their layouts. 
Grid2Im~\cite{Grid2Im}, LostGAN-v2~\cite{LostGAN-v2} and PLGAN~\cite{PLGAN} generate images with distorted and unreal objects.

Especially when input a set of multiple objects with complex relationships, previous work can hardly generate recognizable objects in the position corresponding to layouts. 
For example, in~\cref{fig:visualization_of_comparision_with_SOTA} (a), (c), and (e), the main objects (e.g. train, zebra, bus) in images are poorly generated in previous work, while our LayoutDiffusion generates well.
In~\cref{fig:visualization_of_comparision_with_SOTA} (b), only our LayoutDiffusion generates the laptop in the right place. The images generated by our LayoutDiffusion are more sensorially similar to the real ones.

\begin{figure}[t]
  \centering
   \includegraphics[width=0.90\linewidth]{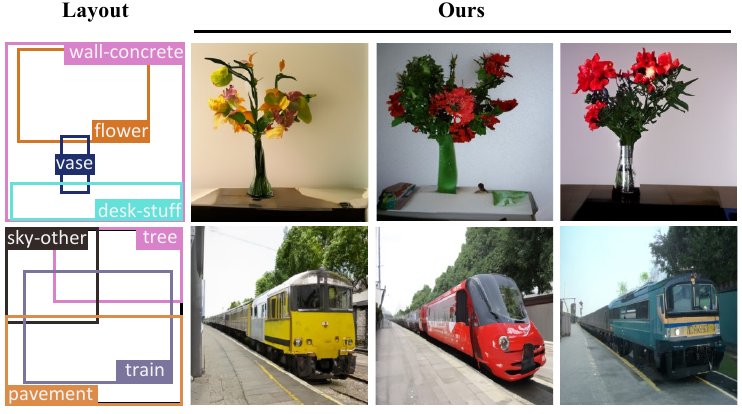} 
   \vspace{-0.5em}
   \caption{The diversity of LayoutDiffusion. 
   Each row of images are from the same layout and have great difference.  
   }
   \label{fig:diversity_of_our_model}
   \vspace{-2.0em}
\end{figure}

We show the diversity of LayoutDiffusion in~\cref{fig:diversity_of_our_model}. 
Images from the same layouts have high quality and diversity (different lighting, textures, colors, and details). 

\begin{figure}[t]
  \centering
   \includegraphics[width=1.0\linewidth]{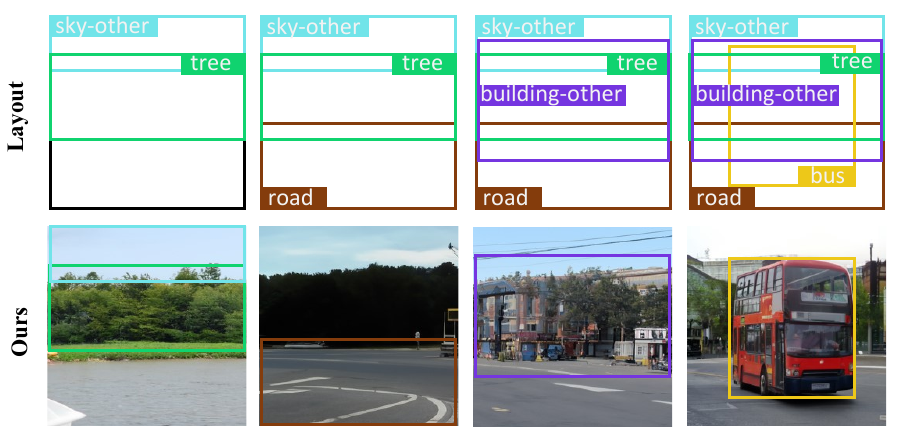}
   \vspace{-0.5em}
   \caption{
   The interactivity of LayoutDiffusion. 
   We add extra layout continuously, and the new objects are also with high quality.
   }
   \label{fig:interactivity}
   \vspace{-1.5em}
\end{figure}

We continuously add an additional layout from the initial layout, the one in the upper left corner, as shown in~\cref{fig:interactivity}.  
In each step, LayoutDiffusion adds the new object in very precise locations with consistent image quality, showing user-friendly interactivity.

\vspace{-0.5em}
\subsection{Quantitative results}
\label{subsec:quantitative_results}
\vspace{-0.5em}
\cref{tab:quantitative_results1} provides the comparison among previous works and our method in FID, IS,  DS, CAS and YOLOScore. Compared to the SOTA method, the proposed method achieves the best performance in comparison.

In overall generation quality, our LayoutDiffusion outperforms the SOTA model by 46.35\% and 29.29\% at most in FID and IS, respectively. 
While maintaining high overall image quality, we also show precise and accurate controllability, LayoutDiffusion outperforms the SOTA model by 122.22\% and 41.82\% at most on YOLOScore and CAS, respectively. 
As for diversity, our LayoutDiffusion still achieves 11.30\% imporvement at most accroding to the DS.
Experiments on these metrics show that our methods can successfully generate the higher-quality images with better location and quantity control.  

In particular, we conduct experiments compared to LDM~\cite{LDM} in~\cref{tab:ablation_study2}. 
``Ours-small'' uses comparable GPU resources to have better FID performance with much fewer parameters and better throughout compared to LDM-8 when ``Ours-small'' outperforms LDM-4 in all respects. 
The results of ``Ours'' indicate that LayoutDiffusion can have better FID performance, 31.6, at a higher cost. 
From these results, LayoutDiffusion always achieves better performance at different cost levels compared with LDM~\cite{LDM}.

\vspace{-0.5em}
\subsection{Ablation studies}
\label{subsec:ablation_studies}
\vspace{-0.4em}
We validate the effectiveness of LFM and OaCA in \cref{tab:ablation_study}, using the evaluation metrics in \cref{subsec:evaluation_metrics}. The significant improvement on FID, IS, CAS, and YOLOScore proves that the application of LFM and OaCA allows for higher generation quality and diversity, along with more controllability.
Furthermore, when applying both, considerable performance, 13.37 / 6.58 / 39.77 / 27.00 on FID / IS / CAS / YOLOScore, is gained.

An interesting phenomenon is that the change of the Diversity Score (DS) is in the opposite direction of other metrics. 
This is because DS, which stands for diversity, is physically the opposite of the controllability represented by other metrics such as CAS and YOLOScore.
The precise control offered on generated image leads to more constraints on diversity. As a result, the Diversity Score (DS) has a slight drop compared to the baseline.




\section{Limitations \& Societal Impacts}
\label{sec:limitations_and_societal_impacts}
\vspace{-0.3em}
\noindent\textbf{Limitations.}
Despite the significant improvements in various metrics, it is still difficult to generate a realistic image with no distortion and overlap, especially for a complex multi-object layout. 
Moreover, the model is trained from scratch in the specific dataset that requires detection labels. How to combine text-guided diffusion models and inherit parameters pre-trained on massive text-image datasets remains a future research.

\noindent\textbf{Societal Impacts.}
Trained on the real-world datasets such as COCO~\cite{COCO-stuff} and VG~\cite{VG}, LayoutDiffusion has the powerful ability to learn the distribution of data and we should pay attention to some potential copyright infringement issues.

\vspace{-0.5em}
\section{Conclusion}
\label{sec:conclusion}
\vspace{-0.5em}
In this paper, we have proposed a one-stage end-to-end diffusion model named LayoutDiffuison, which is novel for the task of layout-to-image generation. 
With the guidance of layout, the diffusion model allows more control over the individual objects while maintaining higher quality than the prevailing GAN-based methods. By constructing a structural image patch with region information, we regrad each patch as a special object and accomplish the difficult multimodal image-layout fusion in a unified form. Specifically, Layout Fusion Module and Object-aware Cross Attention are proposed to model the relationship among multiple objects and fuse the patched image feature with layout at multiple resolutions, respectively.  
Experiments in challenging COCO-stuff and Visual Genome (VG) show that our proposed method significantly outperforms both state-of-the-art GAN-based and diffusion-based methods in various evaluation metrics.

\vspace{1em}
\noindent\textbf{Acknowledgements.}
This work is supported in part by National Natural Science Foundation of China under Grant U20A20222, National Science Foundation for Distinguished Young Scholars under Grant 62225605, National Key Research and Development Program of China under Grant 2020AAA0107400, Research Fund of ARC Lab, Tencent PCG, Zhejiang – Singapore Innovation and AI Joint Research Lab, Ant Group through CCF-Ant Research Fund, and sponsored by CCF-AFSG Research Fund, CAAI-HUAWEI MindSpore Open Fund as well as CCF-Zhipu AI Large Model Fund(CCF-Zhipu202302).


\DeclarePairedDelimiterX{\infdivx}[2]{(}{)}{%
  #1\;\delimsize|\delimsize|\;#2%
}

\newcommand{\kld}[2]{\ensuremath{D_{KL}\infdivx{#1}{#2}}\xspace}

\onecolumn
\appendix

\begin{center}
{\Large \bf Appendix for LayoutDiffusion \par}
\end{center}


\section{More Visualizations}
\label{supplement_subsec:more_visualizations}
\subsection{Controllable Layout Edit by modifying objects in number, position, size, and category}
\label{supplement_subsec:layout_edit}
\begin{figure}[h!]
  \centering
   \includegraphics[width=0.81\linewidth]{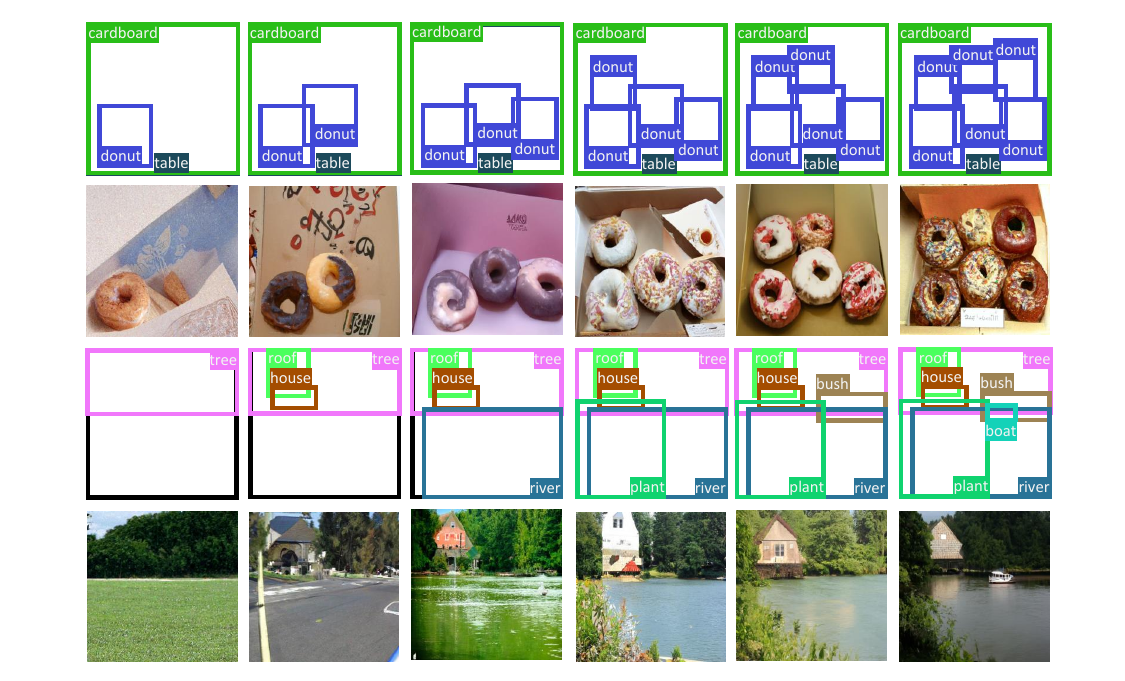}
   \caption{Layout edit by adding objects.}
   \label{fig:layout_edit1}
\end{figure}
\FloatBarrier
\begin{figure}[h!]
  \centering
   \includegraphics[width=0.81\linewidth]{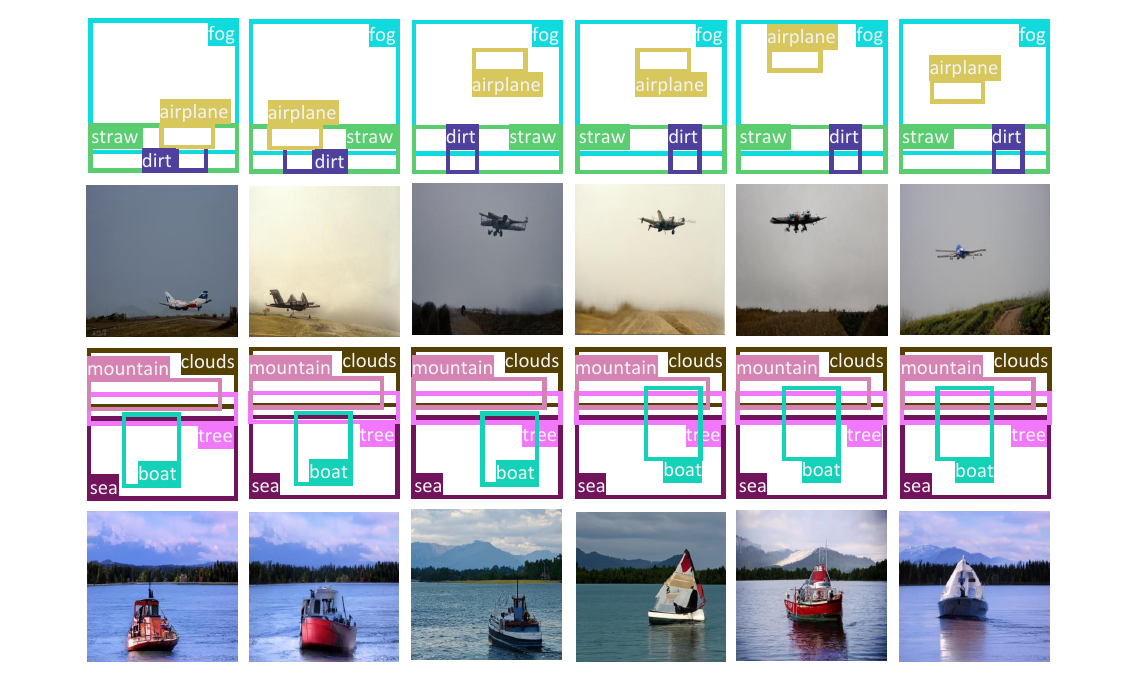}
   \caption{Layout edit by modifying the position of objects.}
   \label{fig:layout_edit2}
\end{figure}
\FloatBarrier

\begin{figure}[h!]
  \centering
   \includegraphics[width=0.95\linewidth]{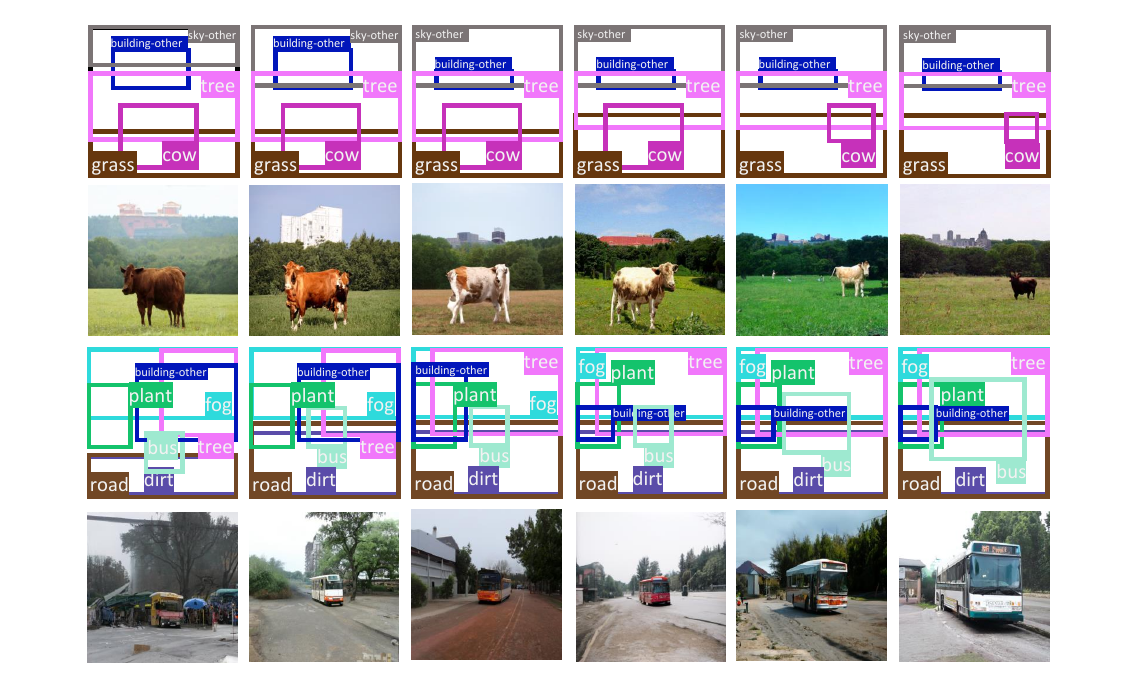}
   \caption{Layout edit by modifying the size of objects.}
   \label{fig:layout_edit3}
\end{figure}
\FloatBarrier
\begin{figure}[h!]
  \centering
   \includegraphics[width=0.95\linewidth]{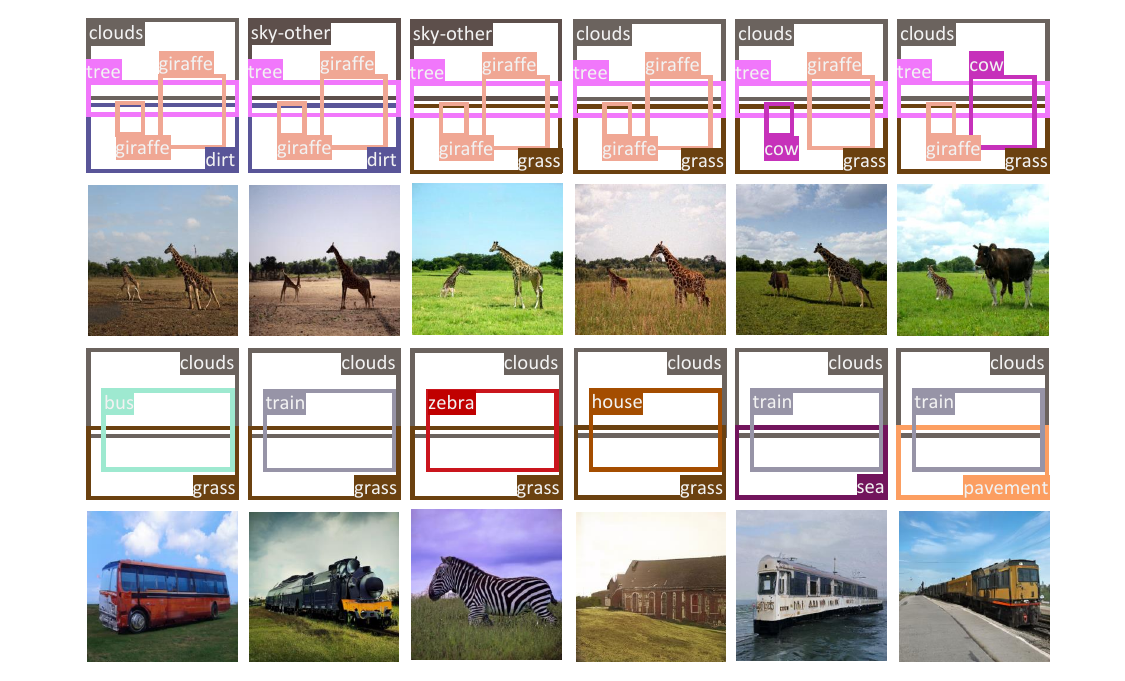}
   \caption{Layout edit by modifying the categories of objects.}
   \label{fig:layout_edit4}
\end{figure}
\FloatBarrier

\clearpage
\subsection{More visualizations on COCO-stuff}
\label{supplement_subsec:more_visualizations_on_coco_stuff_256x256}
\begin{figure}[h!]
  \centering
   \includegraphics[width=0.80\linewidth]{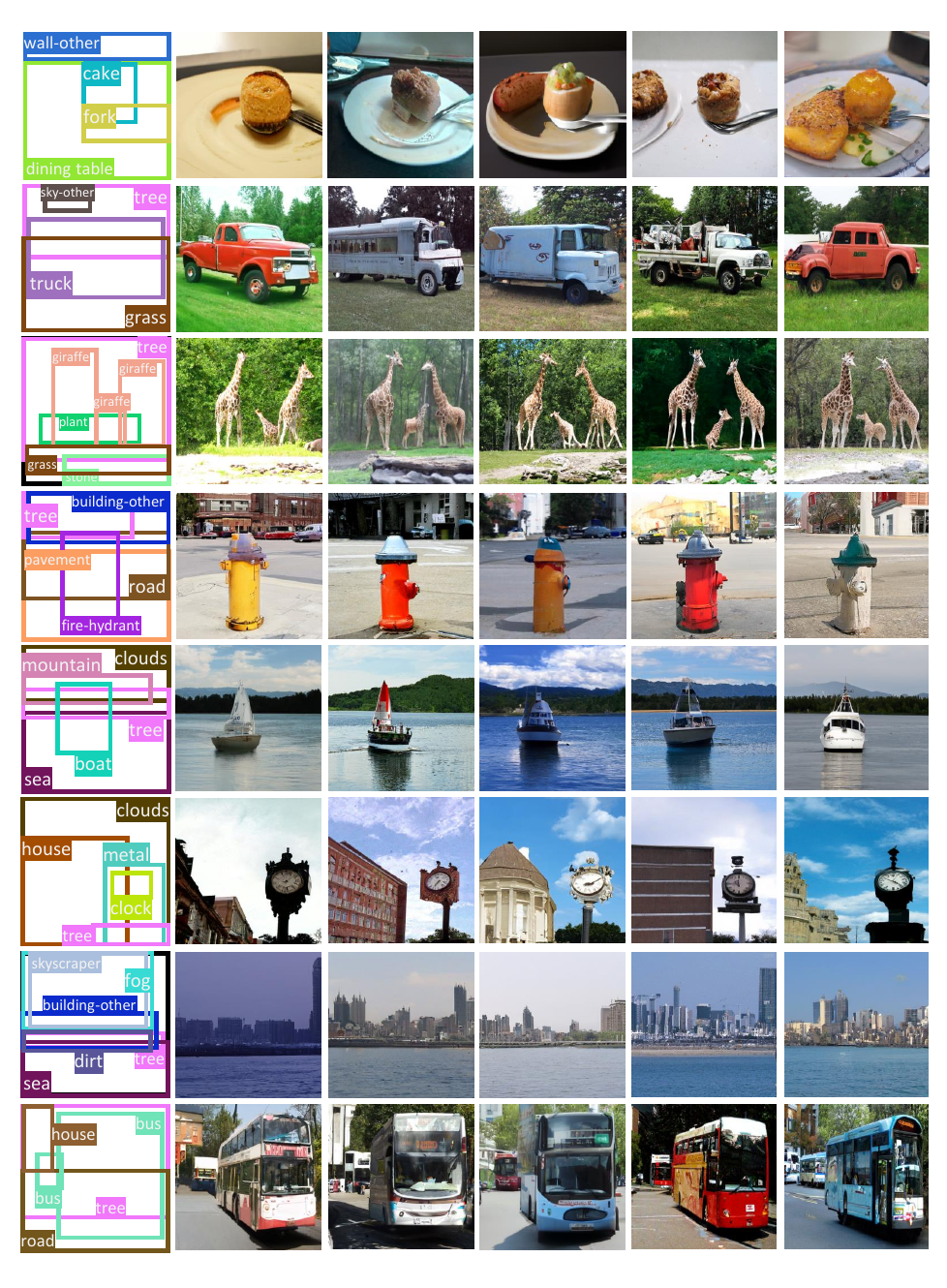}
   \caption{More visualizations on COCO-stuff 256$\times$256. LayoutDiffusion is trained by 1.15M iterations, and sample images using scale=1.0 and dpm-solver 25 steps. The COCO image IDs (from top to bottom) are 85195, 174004, 296969, 338560, 451090, 512248, 573008, 574425.}
   \label{fig:more_visualizations_on_coco_stuff_256x256}
\end{figure}
\FloatBarrier


\subsection{More visualizations on Visual Genome}
\label{supplement_subsec:more_visualizations_on_visual_genome_256x256}
\begin{figure}[h!]
  \centering
   \includegraphics[width=0.84\linewidth]{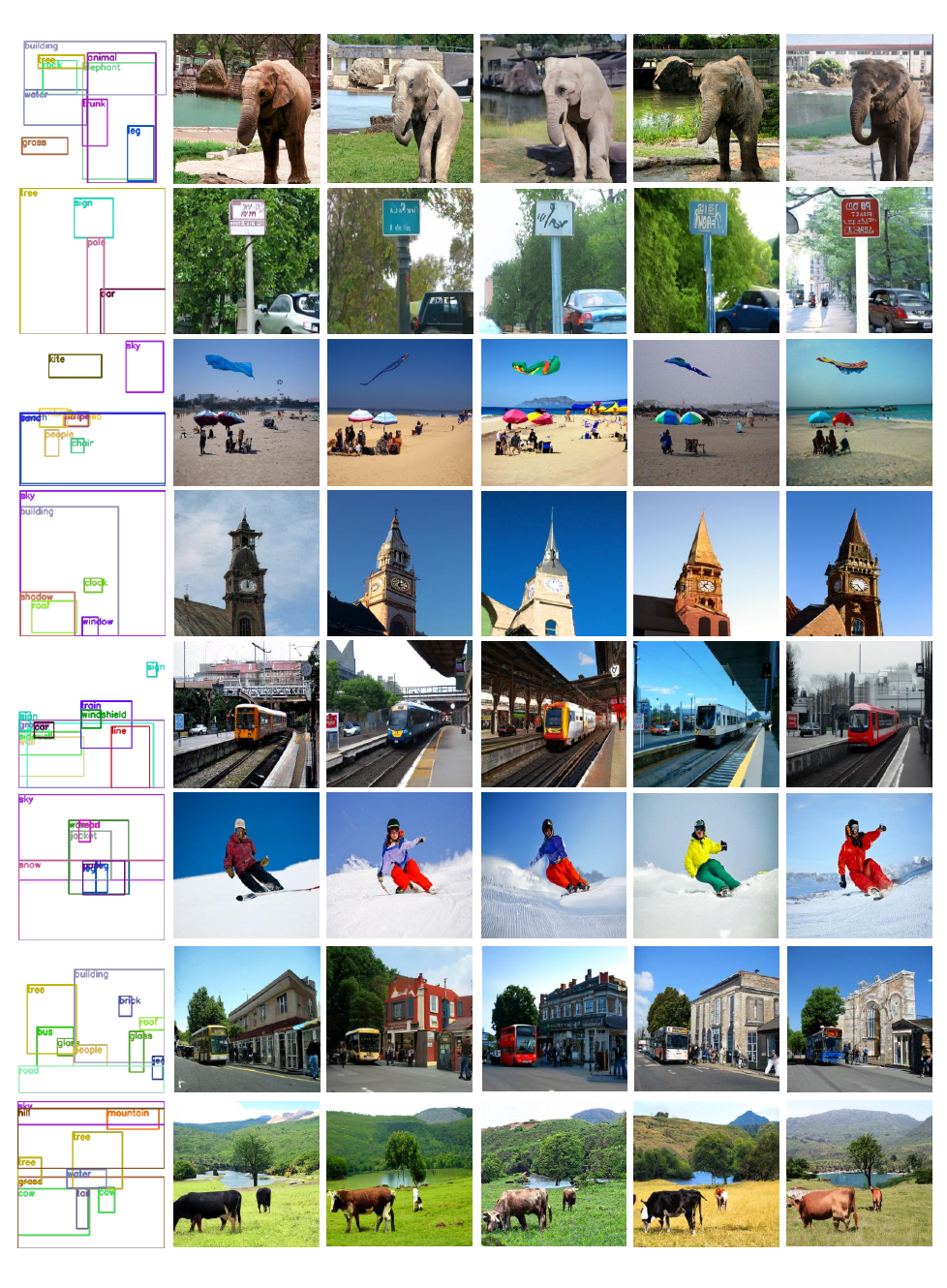}
   \caption{More visualizations on Visual Genome 256$\times$256. LayoutDiffusion is trained by 1.45M iterations, and sample images using scale=1.0 and dpm-solver 25 steps. The VG image IDs (from top to bottom) are 2380568, 2382403, 2382599, 2383021, 2383488, 2385225, 2385290, 2385812.}
   \label{fig:more_visualizations_on_visual_genome_256x256}
\end{figure}
\FloatBarrier


\subsection{More comparision with previous methods}
\label{supplement_subsec:more_comparision_with_previous_methods}

\begin{figure}[h!]
  \centering
   \includegraphics[width=0.84\linewidth]{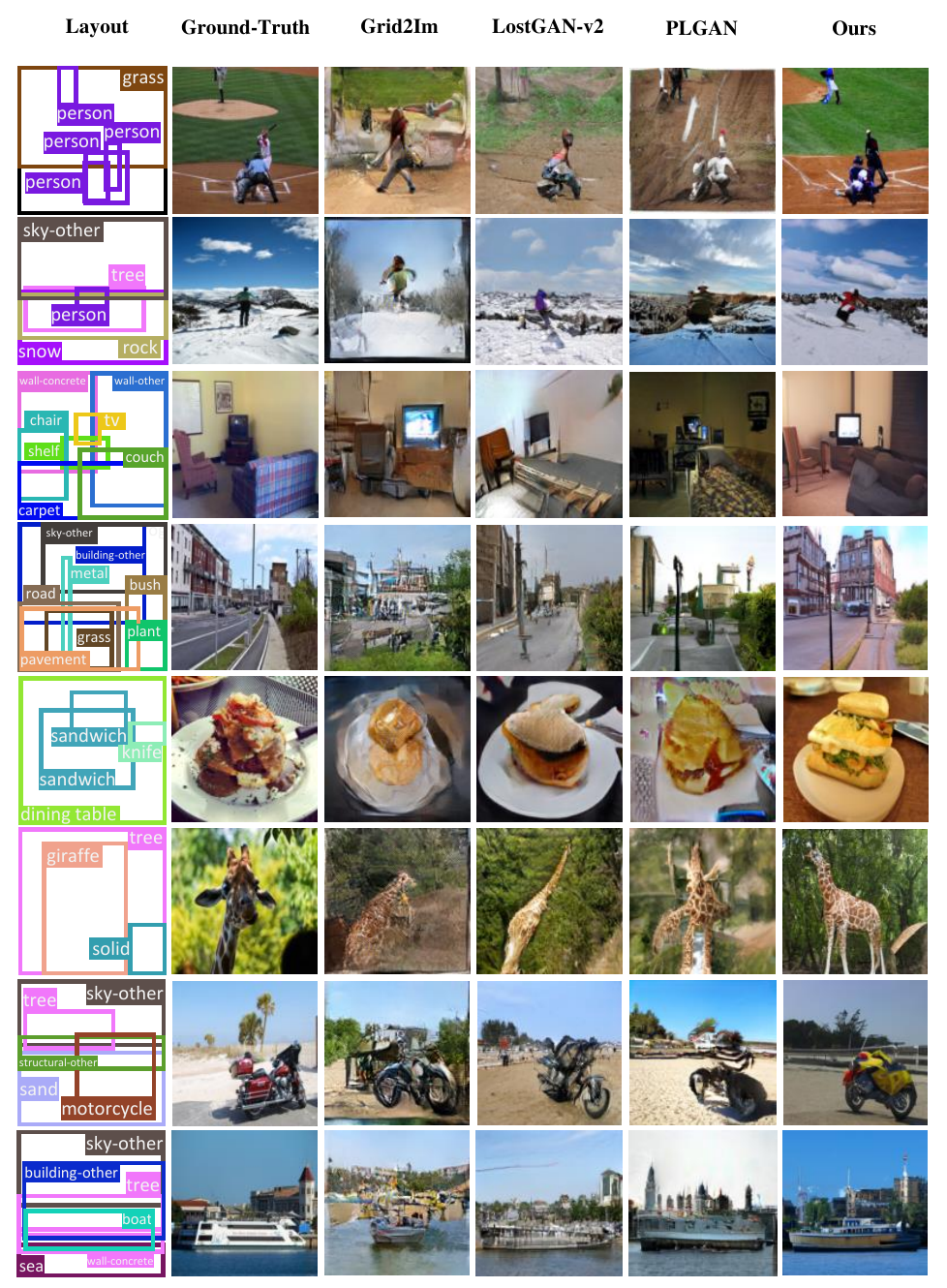}
   \caption{More comparision with previous methods on COCO-stuff 128$\times$128. LayoutDiffusion is trained by 300K iterations, and sample images using scale=0.6 and dpm-solver 25 steps. The COCO image IDs (from top to bottom) are 2153, 2352, 4495, 6723, 10583, 17031, 18737, 543300.}
   \label{fig:more_compare_coco_128}
\end{figure}
\FloatBarrier

\begin{figure}[h!]
  \centering
   \includegraphics[width=0.9\linewidth]{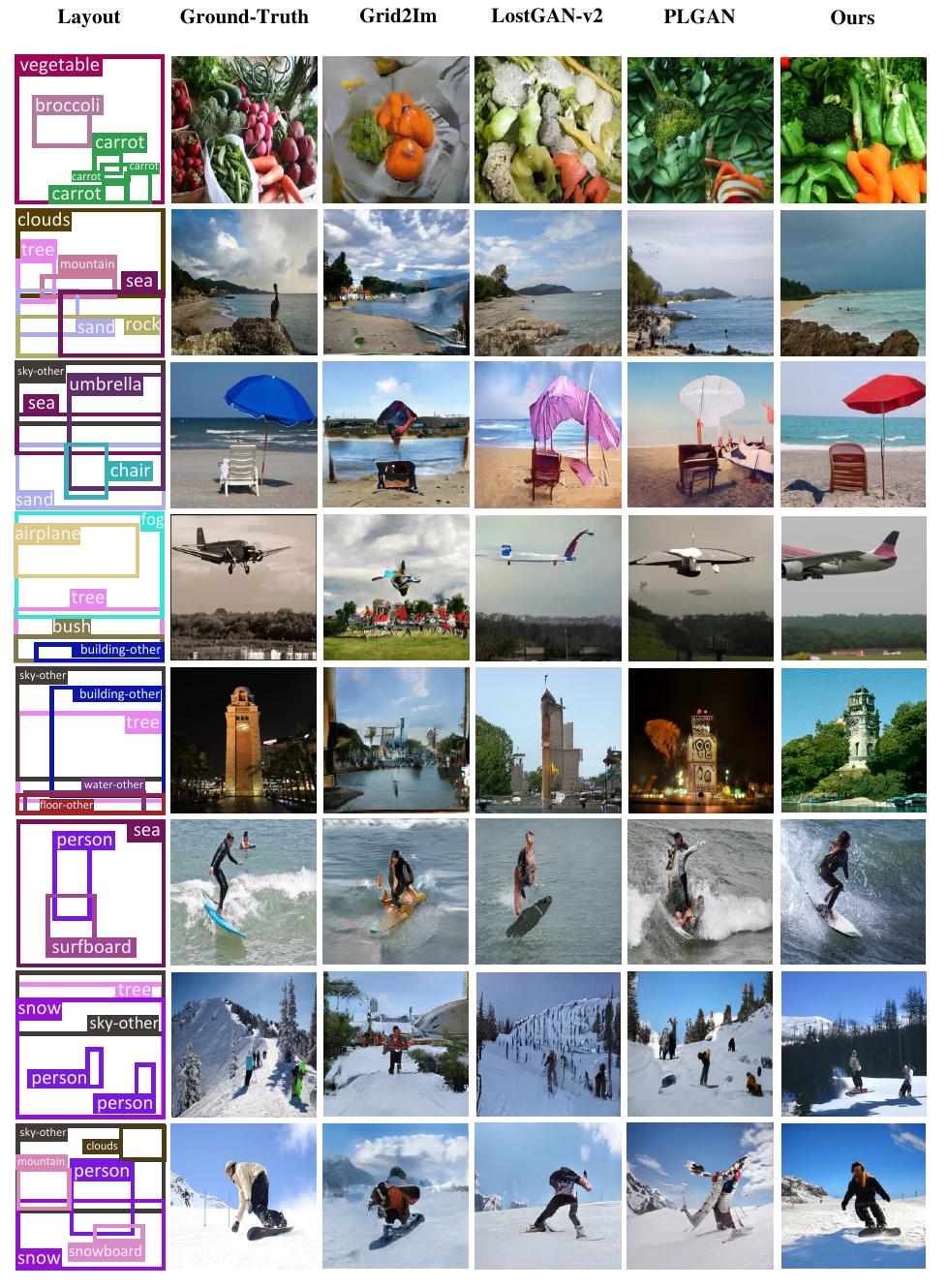}
   \caption{More comparision with previous methods on COCO-stuff 256$\times$256. LayoutDiffusion is trained by 1.15M iterations, and sample images using scale=1.0 and dpm-solver 25 steps. The COCO image IDs (from top to bottom) are 23781, 55299, 84477, 137950, 243034, 252701, 341719, 350405.}
   \label{fig:more_compare_coco_256}
\end{figure}
\FloatBarrier

\begin{figure}[h!]
  \centering
   \includegraphics[width=0.9\linewidth]{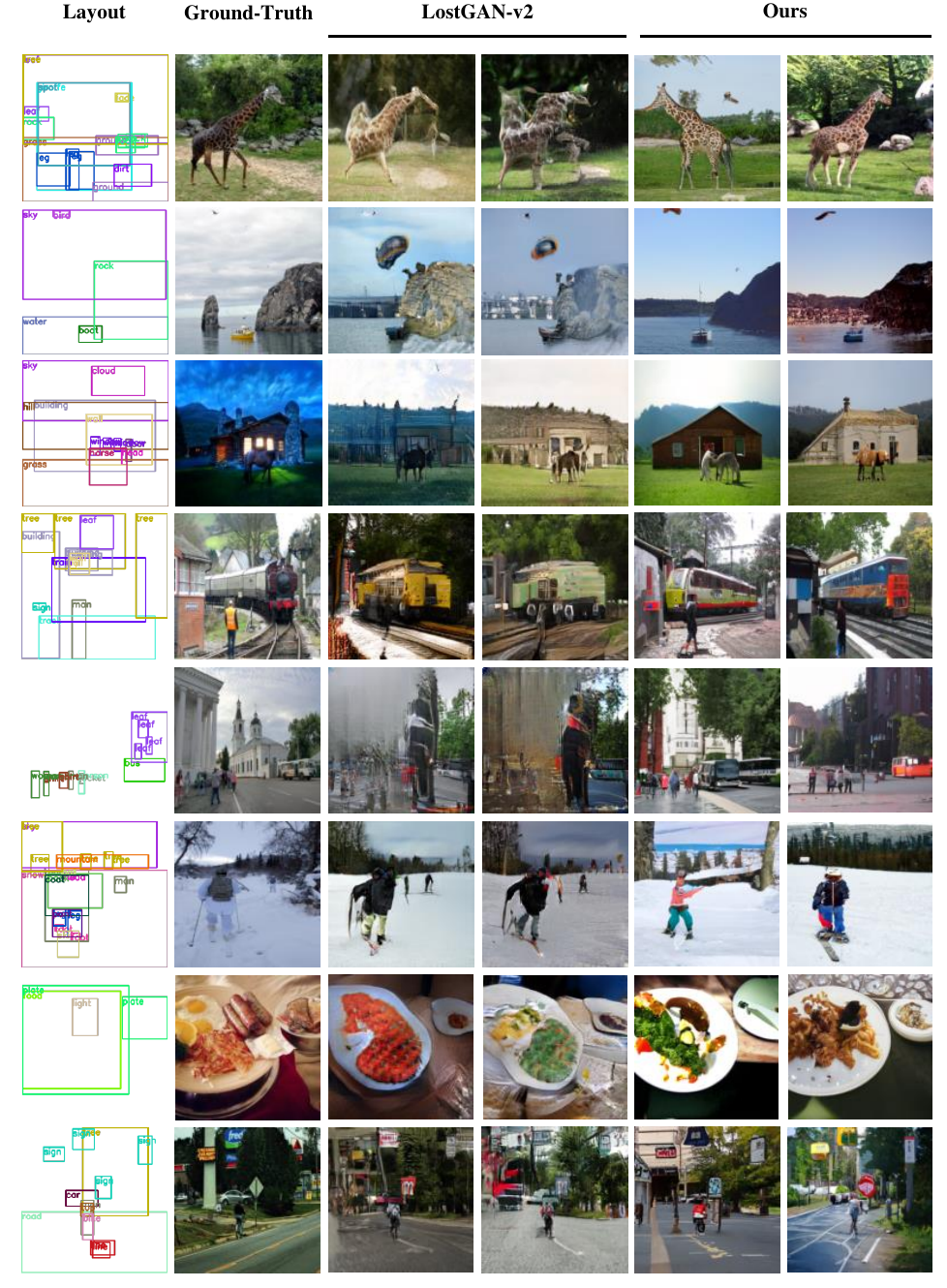}
   \caption{More comparision with previous methods on VG 128$\times$128. LayoutDiffusion is trained by 300K iterations, and sample images using scale=0.5 and dpm-solver 25 steps. The VG image IDs (from top to bottom) are 107945, 150280, 150409, 1160185, 1591817, 1592132, 2341006, 2341475.}
   \label{fig:more_compare_vg_128}
\end{figure}
\FloatBarrier

\begin{figure}[h!]
  \centering
   \includegraphics[width=0.9\linewidth]{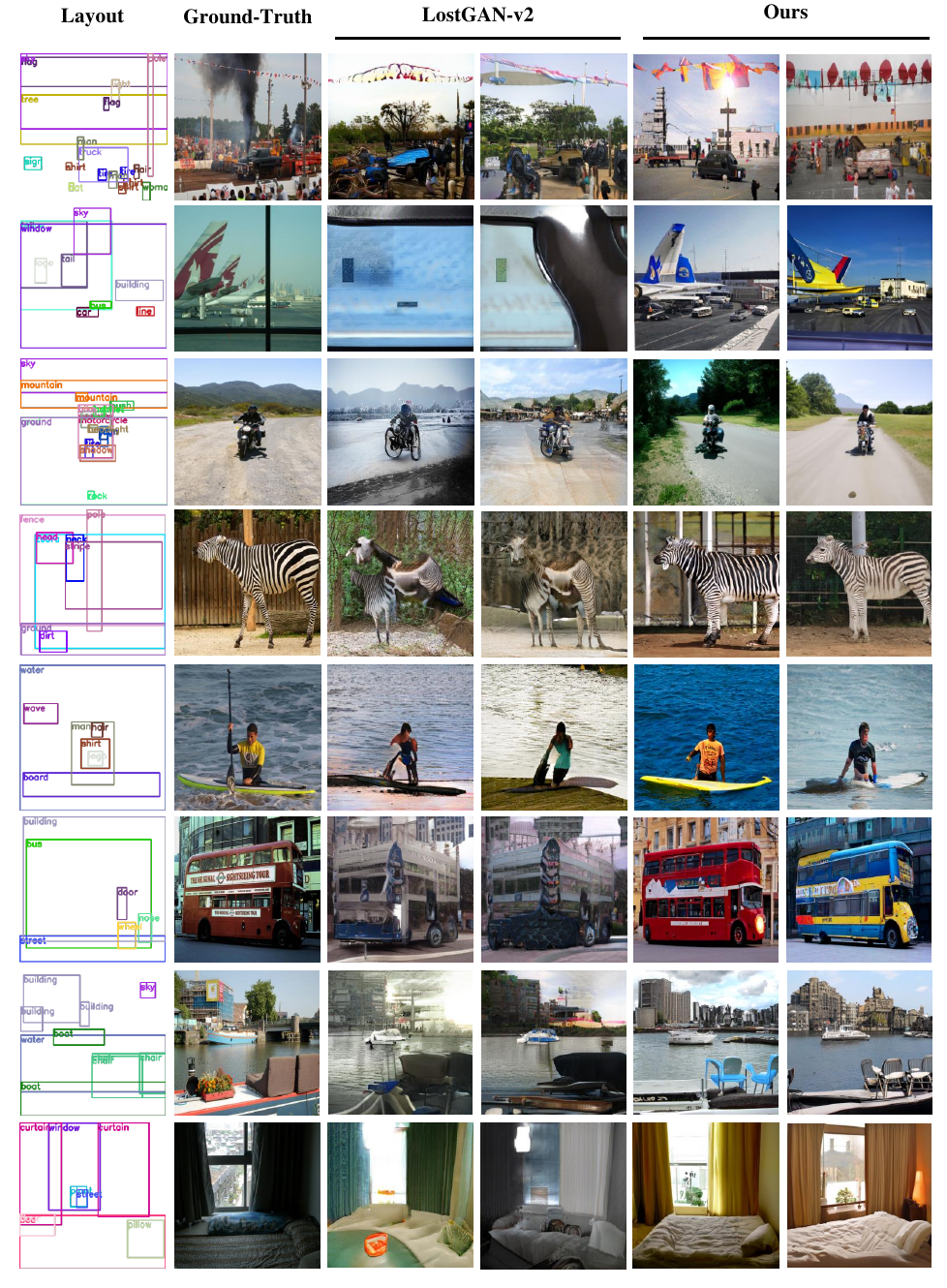}
   \caption{More comparision with previous methods on VG 256$\times$256. LayoutDiffusion is trained by 1.45M iterations, and sample images using scale=1.0 and dpm-solver 25 steps. The VG image IDs (from top to bottom) are 150297, 150358, 1159865, 2340978, 2341300, 2343290, 2344627, 2375966.}
   \label{fig:more_compare_vg_256}
\end{figure}
\FloatBarrier

\newpage
\subsection{Different Scales}
\label{supplement_subsec:different_scales}
\begin{figure}[h!]
  \centering
   \includegraphics[width=0.9\linewidth]{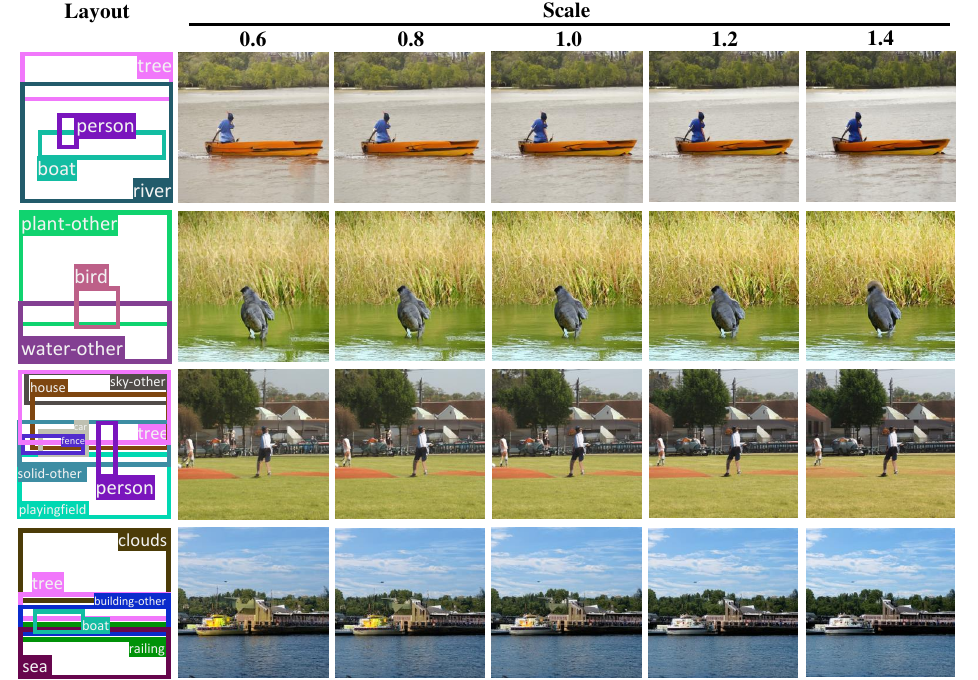}
   \caption{Visualizations on COCO-stuff 256$\times$256 sampled with different scale. LayoutDiffusion is trained by 1.15M iterations, and sample images using dpm-solver 25 steps. The COCO image IDs (from top to bottom) are 472298, 475223, 504415, 513181.}
   \label{fig:different_scales}
\end{figure}
\FloatBarrier

\section{Details of Diffusion Models}
\label{supplement_sec:details_of_diffusion_models}

\subsection{Denoising Diffusion Probabilistic Model (from ADM-G)}
In this section, we will review the formulation of Gaussian diffusion models introduced by DDPM~\cite{DDPM}. A data point is defined as $x_0 \sim q(x_0)$. By gradually adding noise to the clean data $x_0$, we can obtain the noised samples from $x_1$ to $x_T$, where $T$ denotes the maximum steps. Specifically, Gaussian noise according to some variance schedule given by $\beta_t$ is added to $x_{t-1}$ in step $t$ of the Markovian noising process $q$:
\begin{alignat}{2}
q(x_t|x_{t-1}) &=\mathcal{N}(x_t; \sqrt{1-\beta_t}x_{t-1}, \beta_t \mathbf{I})
\end{alignat}

Due to the convenient nature of Gaussian noise, we do not need to apply $t$ times of $q(x_t|x_{t-1})$ repeatedly to sample from $x_t \sim q(x_t|x_0)$. Instead, $q(x_t|x_0)$ can be directly sampled from a Gaussian distribution:
\begin{alignat}{2}
    q(x_t|x_0) &= \mathcal{N}(x_t; \sqrt{\bar{\alpha}_t} x_0, (1-\bar{\alpha}_t) \mathbf{I}) \\
    &= \sqrt{\bar{\alpha}_t} x_0 + \epsilon \sqrt{1-\bar{\alpha}_t},\text{  } \epsilon \sim \mathcal{N}(0, \mathbf{I}), \label{eq:jumpnoise}
\end{alignat}
where $\alpha_t =1 - \beta_t$ and $\bar{\alpha}_t =\prod_{s=0}^{t} \alpha_s$.
The noise variance for an arbitrary timestep is defined as $1 - \bar{\alpha}_t$ in \cref{eq:jumpnoise}, and we could equivalently use this to define the noise schedule instead of $\beta_t$.
DDPM~\cite{DDPM} notes that the posterior $q(x_{t-1}|x_t,x_0)$ is also a Gaussian using with mean $\tilde{\mu}_t(x_t,x_0)$ and variance $\tilde{\beta}_t$, and is defined as follows:
\begin{alignat}{2}
    q(x_{t-1}|x_t,x_0) &= \mathcal{N}(x_{t-1}; \tilde{\mu}(x_t, x_0), \tilde{\beta}_t \mathbf{I}) \label{eq:posterior} \\
    \tilde{\mu}_t(x_t,x_0) &=\frac{\sqrt{\bar{\alpha}_{t-1}}\beta_t}{1-\bar{\alpha}_t}x_0 + \frac{\sqrt{\alpha_t}(1-\bar{\alpha}_{t-1})}{1-\bar{\alpha}_t} x_t \label{eq:mutilde} \\
    \tilde{\beta}_t &=\frac{1-\bar{\alpha}_{t-1}}{1-\bar{\alpha}_t} \beta_t \label{eq:betatilde} 
\end{alignat}

If the total noise added throughout the markov chain is large enough when $T \to \infty$ and correspondingly $\beta_t \to 0$,  the $x_T$ will be well approximated by $\mathcal{N}(0, \mathbf{I})$. This nice property ensures that we can reverse the above forward process and sample from $x_T \sim \mathcal{N}(0, \mathbf{I})$, which is a Gaussian noise. However, since the entire dataset is needed, we cannot easily estimate the posterior $q(x_{t-1} | x_t, x_0)$. Instead, we have to learn a model $p_\theta(x_{t-1} | x_t)$ to approximate it:
\begin{alignat}{2}
p_{\theta}(x_{t-1}|x_t) &=\mathcal{N}(x_{t-1};\mu_{\theta}(x_t, t), \Sigma_{\theta}(x_t, t)) \label{eq:ptheta}
\end{alignat}

To ensure that the estimated $p_{\theta}(x_{t-1}|x_t)$ can learrn the true data distribution $q(x_0)$, we can optimize the following variational lower bound $L_{\text{vlb}}$ for $p_{\theta}(x_0)$:
\begin{alignat}{2}
    L_{\text{vlb}} &=L_0 + L_1 + ... + L_{T-1} + L_T \label{eq:loss} \\
    L_{0} &=-\log p_{\theta}(x_0 | x_1) \label{eq:loss0} \\
    L_{t-1} &=\kld{q(x_{t-1}|x_t,x_0)}{p_{\theta}(x_{t-1}|x_t)} \label{eq:losst} \\
    L_{T} &=\kld{q(x_T | x_0)}{p(x_T)} \label{eq:lossT}
\end{alignat}

Although the above objective is well justified, DDPM~\cite{DDPM} applied a different objective that produces better samples in practice. Specifically, they do not directly predict $\mu_{\theta}(x_t,t)$ as the output of a neural network, but instead train a model $\epsilon_{\theta}(x_t,t)$ to predict $\epsilon$ from Equation \ref{eq:jumpnoise}. This simplified objective is defined as follows:
\begin{alignat}{2}
    L_{\text{simple}} &=E_{t \sim [1,T],x_0 \sim q(x_0), \epsilon \sim \mathcal{N}(0, \mathbf{I})}[||\epsilon - \epsilon_{\theta}(x_t, t)||^2] \label{supplement_eq:lsimple}
\end{alignat}

Then, we can derive $\mu_{\theta}(x_t,t)$ from $\epsilon_{\theta}(x_t,t)$ using the following substitution:
\begin{alignat}{2}
    \mu_{\theta}(x_t,t) &= \frac{1}{\sqrt{\alpha_t}}\left(x_t-\frac{1-\alpha_t}{\sqrt{1-\bar{\alpha}_t}}\epsilon_{\theta}(x_t, t)\right) \label{eq:mufromeps}
\end{alignat}

Note that $\Sigma_{\theta}(x_t,t)$ is not learned in $L_{\text{simple}}$ in DDPM~\cite{DDPM} and is fixed as a constant such as $\beta_t \mathbf{I}$ or $\tilde{\beta}_t \mathbf{I}$, corresponding to upper and lower bounds for the true reverse step variance~\cite{DDIM}, respectively.

\subsection{Classifier-free Method for Layout-conditional Training and Sampling}
Instead of training a separate classifier model, Ho \& Salimans ~\cite{classifier-free} choose to train a diffusion model that allows for both conditional and unconditional sampling, where the unconditional diffusion model $p_{\theta}(x_t, t)$ is parameterized through a score estimator $\epsilon_{\theta}(x_t, t)$ and the conditional diffusion model $p_{\theta}(x_t, t | c)$ is parameterized through $\epsilon_{\theta}(x_t, t, c)$. 

They use a single model to parameterize both conditional and unconditional models, where for the unconditional model they simply input a null token $\varnothing$ for the class identifier $c$ when predicting the score, i.e.\ $\epsilon_{\theta}(x_t, t) = \epsilon_{\theta}(x_t, t, c = \varnothing)$. 
They jointly train this unified model by randomly replacing $c$ with the unconditional class identifier $\varnothing$ with probability $p_\mathrm{uncond}$.
Then the conditional score estimate $\epsilon_\theta(x_t, t, c)$ is replaced with $\tilde{\epsilon}_\theta(x_t, t, c)$ using the following equation:
\begin{align}
    \tilde{\epsilon}_\theta(x_t, t, c) = \epsilon_\theta(x_t, t, c) + s( \epsilon_\theta(x_t, t, c) - \epsilon_{\theta}(x_t, t)) \label{eq:classifier_free_score},
\end{align}
which can be considered as a linear combination of conditional and unconditional score estimates. $s$ is the scale and 
Since \cref{eq:classifier_free_score} has no classifier gradient, no more gradient calculation in classifier guidance is needed during sampling. 
Furthermore, $s$ can be changed to modify the effect of the condition.

In the layout-to-image generation, $c$ is the layout $l$ defined in Sec. 3.1. Layout Embedding and $\varnothing$ is the emtpy layout $l_{\text{pad}}$ defined in Sec. 3.4. Layout-conditional Diffusion Model.

\clearpage
\section{Implementation details}
\label{supplement_sec:implementation_details}

\subsection{ Hyperparameters}
\label{supplement_sec:experiments}

\begin{table}[h!]
\begin{center}
    \begin{adjustbox}{max width=1.0\textwidth}
        \begin{tabular}{l cc ccc}
            \toprule
                \textbf{Dataset} & \multicolumn{2}{c}{COCO-stuff 256$\times$256} & COCO-stuff 128$\times$128 & VG 256$\times$256 & VG 128$\times$128\\
                \cmidrule(lr){2-3} \cmidrule(lr){4-4} \cmidrule(lr){5-5} \cmidrule(lr){6-6} 
                \textbf{Model} & LayoutDiffusion & LayoutDiffusion-small & LayoutDiffusion  & LayoutDiffusion & LayoutDiffusion   \\
            \midrule
                \rowcolor{gray!20}Layout-conditional Diffusion Model & & & & & \\
                In Channels                &    3     &    3     &    3      &    3     &    3     \\
                Out Channels               &    6     &    6     &    6      &    6     &    6     \\
                Hidden Channels             &   256    &   128    &   256     &   256    &   256    \\
                Channel Multiply           & 1,1,2,2,4,4 & 1,1,2,2,4,4 & 1,1,2,3,4 & 1,1,2,2,4,4 & 1,1,2,3,4 \\
                Number of Residual Blocks  &    2     &    2     &    2      &    2     &    2     \\
                Dropout & 0 & 0 & 0.1 & 0.1 & 0.1 \\
                Diffusion Steps & 1000 & 1000 & 1000 & 1000 & 1000 \\
                Noise Schedule & linear & linear & linear & linear & linear \\
            \midrule
                \rowcolor{gray!20}Layout-Image Fusion Module & & & & & \\
                Downsampling Scale For Fusion  & 8,16,32 & 8,16,32 & 4,8,16 & 8,16,32 & 4,8,16 \\
                Resolution for Fusion & 32,16,8 & 32,16,8 & 32,16,8 & 32,16,8 & 32,16,8 \\
                Fusion Method & OaCA & OaCA & OaCA & OaCA & OaCA \\
                Number of Attention Blocks &    1     &    1     &    1      &    1     &    1     \\
                Number of Heads & 4 & 4 & 4 & 4 & 4 \\
            \midrule
                \rowcolor{gray!20}Layout Fusion Module & & & & & \\
                Hidden Channels & 256 & 128 & 256 & 256 & 256 \\
                Transformer Depth & 6 & 4 & 6 & 6 & 6 \\
                Attention Method & Self-Attention & Self-Attention & Self-Attention & Self-Attention & Self-Attention \\
                Number of Heads & 8 & 8 & 8 & 8 & 8 \\
            \midrule
                \rowcolor{gray!20}Layout Embedding & & & & & \\
                Embedding Dimension & 256 & 128 & 256 & 256 & 256 \\
                Maximum Number of Objects & 8 & 8 & 8 & 10 & 10 \\
                Maximum Number of Length & 10 & 10 & 10 & 12 & 12 \\
                Maximum Number of Class Id & 185 & 185 & 185 & 180 & 180 \\
            \midrule
                \rowcolor{gray!20}Training Hyperparameters & & & & & \\
                Total Batch Size & 32 & 32 &  64 & 32 & 64 \\
                Number of GPUs & 8 & 8 & 8 & 8 & 8 \\
                Learning Rate & 1e-5 & 1e-5 & 2e-5 & 1e-5 & 2e-5 \\
                Mixed Precision Training   &   Yes   &   Yes   &   Yes    &   Yes   &   Yes   \\
                Weight Decay & 0 & 0 & 0 & 0 & 0 \\
                EMA Rate & 0.9999 & 0.9999 & 0.9999 & 0.9999 & 0.9999 \\
                Classifier-free Dropout & 0.2 & 0.2 & 0.2 & 0.2 & 0.2 \\
                Iterations & 1.15M & 1.4M & 300K & 1.45M & 300K \\
            \bottomrule
        \end{tabular}
    \end{adjustbox}
\end{center}
\caption{ Hyperparameters for the proposed LayoutDiffusion in Sec. 4.4. Quantitative results. All trained on eight RTX 3090.}
\label{tab:hyperparams_for_LayoutDiffusion}
\end{table}
\FloatBarrier

\begin{table}[h!]
\begin{center}
    \resizebox{0.85 \textwidth}{!}{
        \begin{tabular}{l cc ccc}
            \toprule
                \textbf{Model}                                  & LayoutDiffusion  & $-$ LFM  & $-$ OaCA & $- $OaCA, $+$ CA & $-$ LFM, $-$ OaCA   \\
            \midrule
                Fusion Method                                   &  OaCA   &  OaCA   &    -    &    CA   &    -    \\
                Transformer Depth                               &    6    &   0     &    6    &    6    &    0    \\
                Iterations                                      &  300K   &  300K   &  300K   &  300K   &   300K  \\
            \midrule
                FID                                             &  16.57  & + 0.19 & + 0.49 & - 0.11 & + 13.37  \\
                DS                                              &  0.47   & + 0.01 & + 0.05 & + 0.01 & + 0.23   \\
                CAS                                             &  43.60  & - 2.93 & - 12.74& - 1.13 & - 39.77  \\
                YOLOScore                                       &  27.00  & - 8.20 & - 20.10& - 3.4  & - 27.0  \\
            \bottomrule
        \end{tabular}
    }
\end{center}
\caption{Hyperparameters for the ablation study in Sec. 4.5 . }
\label{tab:hyperparams_for_ablation_study} 
\end{table}
\FloatBarrier

\subsection{Analysis of Training Resources and Sampling Speed}
\label{supplement_sec:analysis_of_training_resources_and_sampling_speed}

\begin{table}[h!]
  \centering
  \resizebox{0.95 \textwidth}{!}{
      \begin{tabular}{l|ccccccc}
        \toprule
        
         \multirow{2}{*}{method} & FID $\downarrow$ & $N_{\text{parms}}$  & Throughout    & Compression stage   & Diffusion stage   & Total \\
                                 &                  &                     &  images / s   & V100 days           & V100 
    days         & V100 days \\ 
        \midrule
         LDM-8  (100 steps) &  42.06  &  345M   &  0.457  &   66    &    3.69    & 69.69   \\
         LDM-4  (200 steps)             &  40.91  &  306M   &  0.267  &   29    &   95.49    & 124.49  \\
         LayoutDiffusion-small (25 steps)                  &  36.16  &  142M   &  0.608  &    -    &   75.83    & 75.83   \\ 
         LayoutDiffusion  (25 steps)                       &  31.68  &  569M   &  0.308  &    -    &   216.55   & 216.55  \\
        \bottomrule
      \end{tabular}
  }
  \caption{Comparison  with SOTA diffusion-based methods LDM on COCO-stuff 256$\times$256. We generate the same 2048 images of LDM for a fair comparision. LDM is sampled using DDIM~\cite{DDIM} and LayoutDiffusion is sampled with DPM-Solver~\cite{DPM-Solver}. The hyperparameters of LayoutDiffusion and LayoutDiffusion-small are listed in ~\cref{tab:hyperparams_for_LayoutDiffusion}}
  \label{tab:analysis_of_training_resources_and_sampling_speed}
\end{table}
\FloatBarrier

\section{Evaluation }
\label{supplement_subsec:evaluation}

\subsection{Datasets}
\textbf{COCO-Stuff}~\cite{COCO-stuff}.
COCO\cite{coco} is a large-scale object detection, segmentation, and captioning dataset. 
COCO-Stuff\cite{COCO-stuff} augments all 164K images of the COCO\cite{coco} dataset with pixel-level stuff annotations. 
These annotations can be used for scene understanding tasks like semantic segmentation, object detection and image captioning. 
COCO-Stuff\cite{COCO-stuff} contains 80 categories of thing and 91 categories of stuff, respectively. 
Following the settings of LostGAN-v2\cite{LostGAN-v2}, we use the COCO 2017 Stuff Segmentation Challenge subset containing 40K / 5k / 5k images for train / val / test-dev set. 
Segmengtation annotation is not used.
We use images in the train and val set with 3 to 8 objects that cover more than 2\% of the image and not belong to ’crowd’. 
Finally, there are 25,210 train and 3,097 val images. 
Some previous works\cite{sg2im, Grid2Im} don't filter objects belong to ’crowd’, causing different numbers of images. Specificly, there are 24,972 train and 3,074 val images. 
LAMA\cite{YOLOScore} uses the full COCO-Stuff\cite{COCO-stuff} 2017 dataset, leaving 74,777 train and 3,097 val images.
\cref{tab:coco_number} summarizes the difference in the number of images by different filtering methods.

\begin{table*}[h!]
  \centering
  \resizebox{0.9 \textwidth}{!}{
      \begin{tabular}{@{}cc|cccccc@{}}
        \toprule
        \textbf{use} & \textbf{filter} & \multicolumn{3}{c}{train} & \multicolumn{3}{c}{val} \\
        \cmidrule(lr){3-5}  \cmidrule(lr){6-8}
        \textbf{deprecated} & \textbf{'crowd'} & image & object & object/image & image & object & object/image\\ 
        \midrule
    
                      &               & 74,121 & 411,682 & 5.55 & 3,074 & 17,100 & 5.56\\
        \checkmark    &               & 24,972 & 138,162 & 5.53 & 3,074 & 17,100 & 5.56\\
                      &   \checkmark  & 74,777 & 414,443 & 5.54 & 3,097 & 17,191 & 5.55\\
        \checkmark    &   \checkmark  & 25,210 & 139,175 & 5.52 & 3,097 & 17,191 & 5.55\\
    
        \bottomrule
      \end{tabular}
  }
  \caption{
  The difference in the number of images by different filtering methods, \textbf{use deprecated} means use COCO 2017 Stuff Segmentation Challenge subset, \textbf{filter 'crowd'} means filter objects belong to ’crowd’.
  }
  \label{tab:coco_number}
\end{table*}
\FloatBarrier

\textbf{Visual Genome}~\cite{VG}.
Following the settings of Sg2Im\cite{sg2im}, we experiment on Visual Genome\cite{VG} version 1.4 (VG) which comprises 108,077 images annotated with scene graphs. Visual Genome\cite{VG} collects images with dense annotations of objects, attributes, and relationships. Here, we only use bounding boxes. We divide the data into 80\% / 10\% / 10\% for the train / val / test set. We select the object / relationship categories occurring at least 2000 / 500 times in the train set, respectively, and select the images with 3 to 30 bounding boxes and ignoring all small objects. Finally, the train / val / test set has 62,565 / 5,062 / 5,096 images.

\subsection{Evaluation Metrics}
\label{supplement_subsec:evaluation_metrics}

Comprehensive evaluations of generated images remains a challenge. We use six metrics, from image-level to layout-level, to evaluate the quality of the generated images and the layout control from different aspects.\\

\textbf{Fr`echet Inception Distance  (FID)}\cite{FID}
shows the difference between the real images and the generated images by using an ImageNet-pretrained Inception-V3\cite{Inceptionv3} network and computing the Fr`echet distance between two Gaussian distributions fitted to generated images and real images respectively. We save the GT images as real images when sampling the generated images. The real images and the generated images are saved to two folder respectively. Then compute the FID score, based on the official code of FID\footnote{the official code of FID\cite{FID}:\url{https://github.com/bioinf-jku/TTUR}}. For FID, the lower the score, the smaller the difference between the generated images and the real images, meaning the generator model is better.\\

\textbf{Inception Score (IS)}\cite{IS}
shows the overall quality of the generated images by using an Inception-V3\cite{Inceptionv3} network pretrained on the ImageNet-1000 classification benchmark and computing a score(statistics) of the network’s outputs with generated images of a generator model. IS measures the quality of images on two aspects: clarity and diversity. We only need the generated images to compute the IS score, based on the official code of IS\footnote{the official code of IS\cite{IS}:\url{https://github.com/openai/improved-gan}}. For IS, the higher the score, the better the quality of generated images, meaning the generator model is better.\\

\textbf{Diversity Score (DS)}
measures the diversity between the generated images from the same layout by comparing the perceptual similarity in a DNN feature space between them. Here, we adopt the LPIPS\cite{LPIPS} metric. For each sample, we repeat two times, use the two images from the same layout to compute the DS score, then calculate mean and std of these scores as the reported DS score, based on the official code of DS\footnote{the official code of DS~\cite{LPIPS}:\url{https://github.com/richzhang/PerceptualSimilarity}}. For DS, the higher the score, the better the diversity between the generated images, meaning the generator model is better.\\

\textbf{YOLO Score}\cite{YOLOScore}
uses a pretrained YOLOv4\cite{YOLOv4} model to evaluate bbox mAP on 80 thing categories based on the official code of LAMA\footnote{the official code of LAMA~\cite{YOLOScore}:\url{https://github.com/ZejianLi/LAMA}} and the official code of YOLOv4\footnote{the official code of YOLOv4~\cite{YOLOv4}:\url{https://github.com/AlexeyAB/darknet}}. YOLO Score\cite{YOLOScore} is proposed to evaluate the alignmentand fidelity of generated objects, measuring how generated objects are recognizable when even the layout is unknow. YOLO\cite{yolo, YOLOv4} is a well-known series of object detector, inferring layouts from the given images. Before send images to detector, they are upsampled to 512$\times$512. And different from LAMA\cite{YOLOScore}, since we filter the objects and images in datasets, we think it is better to evaluate bbox mAP only on filtered annotations.\\

\textbf{Classification Score (CAS)}\cite{CAS}
measures classification accuracy of layout areas on generated images. We crop the GT box area of images and resize objects at a resolution of 32$\times$32 with their class. Then train a ResNet101\cite{ResNet} classifier with cropped images on generated images and test it on cropped images on real images, based on a widely used codebase of image classification\footnote{the widely used codebase of image classification: ~\url{https://github.com/hysts/pytorch_image_classification}}. For CAS, the higher the score, the better the quality of layout control, meaning the generator model is better.

\clearpage

{\small
\bibliographystyle{ieee_fullname}
\bibliography{egbib}
}

\end{document}